\documentclass[10pt,twocolumn,letterpaper]{article}

\usepackage{cvpr}              

\usepackage{graphicx}
\usepackage{amsmath}
\usepackage{amssymb}
\usepackage{booktabs}

\usepackage{courier}  
\usepackage[hyphens]{url}  
\usepackage{graphicx} 
\usepackage{multirow}
\usepackage{multicol}
\usepackage{xcolor}
\usepackage{multicol}
\usepackage{xcolor}
\usepackage{amsmath}
\usepackage{enumitem}
\usepackage{colortbl}
\usepackage{amsmath}
\usepackage{enumitem}

\usepackage[pagebackref,breaklinks,colorlinks]{hyperref}

\newcommand{\egno}{\textit{e}.\textit{g}.} 

\newcommand{\ieno}{\textit{i}.\textit{e}.} 

\newcommand{\etcno}{\textit{etc}} 
\newcommand{\ours}{DFF}
\newcommand{\ourfull}{Deep Frequency Filtering}

\usepackage[capitalize]{cleveref}
\crefname{section}{Sec.}{Secs.}
\Crefname{section}{Section}{Sections}
\Crefname{table}{Table}{Tables}
\crefname{table}{Tab.}{Tabs.}

\def\cvprPaperID{6351} 
\def\confName{CVPR}
\def\confYear{2023}

\begin{document}

\title{Deep Frequency Filtering for Domain Generalization}

\author{
Shiqi Lin$^{1}$\footnotemark[1]\quad  
Zhizheng Zhang$^{2}$\quad  
Zhipeng Huang$^1$\footnotemark[1]\quad 
Yan Lu$^2$\quad 
Cuiling Lan$^2$\quad 
Peng Chu$^2$\\
Quanzeng You$^2$\quad
Jiang Wang$^2$\quad 
Zicheng Liu$^2$\quad
Amey Parulkar$^2$\quad
Viraj Navkal$^2$\quad 
Zhibo Chen$^1$\\
$^1$University of Science and Technology of China \quad
$^2$Microsoft \\
{\tt\small \{linsq047,hzp1104\}@mail.ustc.edu.cn\quad chenzhibo@ustc.edu.cn}\\ 
{\tt\small \{zhizzhang,yanlu,culan,pengchu,quyou,jiangwang,zliu,amey.parulkar,vnavkal\}@microsoft.com} 
}

\maketitle

\renewcommand{\thefootnote}{\fnsymbol{footnote}}
\footnotetext[1]{This work was done when Shiqi Lin and Zhipeng Huang were interns at Microsoft Research Asia.}


\begin{abstract}
Improving the generalization ability of Deep Neural Networks (DNNs) is critical for their practical uses, which has been a longstanding challenge. Some theoretical studies have uncovered that DNNs have preferences for some frequency components in the learning process and indicated that this may affect the robustness of learned features. In this paper, we propose Deep Frequency Filtering (DFF) for learning domain-generalizable features, which is the first endeavour to explicitly modulate the frequency components of different transfer difficulties across domains in the latent space during training. To achieve this, we perform Fast Fourier Transform (FFT) for the feature maps at different layers, then adopt a light-weight module to learn attention masks from the frequency representations after FFT to enhance transferable components while suppressing the components not conducive to generalization. Further, we empirically compare the effectiveness of adopting different types of attention designs for implementing DFF. Extensive experiments demonstrate the effectiveness of our proposed DFF and show that applying our DFF on a plain baseline outperforms the state-of-the-art methods on different domain generalization tasks, including close-set classification and open-set retrieval.
\end{abstract}

\section{Introduction}

Domain Generalization (DG) seeks to break through the \emph{i.i.d.} assumption that training and testing data are identically and independently distributed. This assumption does not always hold in reality since domain gaps are commonly seen between the training and testing data. However, collecting enough training data from all possible domains is costly and even impossible in some practical environments. Thus, learning generalizable feature representations is of high practical value for both industry and academia.

Recently, a series of research works \cite{yin2019fourier} analyze deep learning from the frequency perspective. These works, represented by the F-Principle \cite{xu2019training}, uncover that there are different preference degrees of DNNs for the information of different frequencies in their learning processes. Specifically, DNNs optimized with stochastic gradient-based methods tend to capture low-frequency components of the training data with a higher priority \cite{xu2018understanding} while exploiting high-frequency components to trade the robustness (on unseen domains) for the accuracy (on seen domains) \cite{wang2020high}. This observation indicates that different frequency components are of different transferability across domains.

In this work, we seek to learn generalizable features from a frequency perspective. To achieve this, we conceptualize \ourfull~(\ours), which is a new technique capable of enhancing the transferable frequency components and suppressing the ones not conducive to generalization in the latent space. With \ours, the frequency components of different cross-domain transferability are dynamically modulated in an end-to-end manner during training. This is conceptually simple, easy to implement, yet remarkably effective. In particular, for a given intermediate feature, we apply Fast Fourier Transform (FFT) along its spatial dimensions to obtain the corresponding frequency representations where different spatial locations correspond to different frequency components. In such a frequency domain, we are allowed to learn a spatial attention map and multiply it with the frequency representations to filter out the components adverse to the generalization across domains. 

The attention map above is learned in an end-to-end manner using a lightweight module, which is instance-adaptive. As indicated in \cite{xu2018understanding,wang2020high}, low-frequency components are relatively easier to be generalized than high-frequency ones while high-frequency components are commonly exploited to trade robustness for accuracy. Although this phenomenon can be observed consistently over different instances, it does not mean that high-frequency components have the same proportion in different samples or have the same degree of effects on the generalization ability. Thus, we experimentally compare the effectiveness of task-wise filtering with that of instance-adaptive filtering. Here, the task-wise filtering uses a shared mask over all instances while the instance-adaptive filtering uses unshared masks. We find the former one also works but is inferior to our proposed design by a clear margin. As analyzed in \cite{chi2020fast}, the spectral transform theory \cite{katznelson2004introduction} shows that updating a single value in the frequency domain globally affects all original data before FFT, rendering frequency representation as a global feature complementary to the local features learned through regular convolutions. Thus, a two-branch architecture named Fast Fourier Convolution (FFC) is introduced in \cite{katznelson2004introduction} to exploit the complementarity of features in the frequency 
and original domains with an efficient ensemble. To evaluate the effectiveness of our proposed \ours, we choose this two-branch architecture as a base architecture and apply our proposed frequency filtering mechanism to its spectral transform branch. Note that FFC provides an effective implementation for frequency-space convolution while we introduce a novel frequency-space attention mechanism. We evaluate and demonstrate our effectiveness on top of it. 

Our contributions can be summarized in the following:
\begin{itemize}[noitemsep,nolistsep,leftmargin=*]
\item We discover that the cross-domain generalization ability of DNNs can be significantly enhanced by a simple learnable filtering operation in the frequency domain.
\item We propose an effective Deep Frequency Filtering (\ours) module where we learn an instance-adaptive spatial mask to dynamically modulate different frequency components during training for learning generalizable features. 
\item We conduct an empirical study for the comparison of different design choices on implementing \ours, and find that the instance-level adaptability is required when learning frequency-space filtering for domain generalization. 
\end{itemize}

\section{Related Work}

\subsection{Domain Generalization}

Domain Generalization (DG) aims to improve the generalization ability of DNNs from source domains to unseen domains, which is widely needed in different application scenarios. The challenges of DG have been addressed from \emph{data}, \emph{model}, and \emph{optimization} perspectives. From the \emph{data} perspective, augmentation \cite{yue2019domain,honarvar2020domain,tremblay2018training} and generation \cite{shankar2018generalizing,volpi2018generalizing,zhou2020deep} technologies are devised to increase the diversity of training samples so as to facilitate generalization. From the \emph{model} perspective, some efforts are made to enhance the generalization ability by carefully devising the normalization operations in DNNs \cite{pan2018two,liu2022debiased,seo2020learning} or adopting an ensemble of multiple expert models \cite{zhou2021domain,mancini2018best}. From the \emph{optimization} perspective, there are many works designing different training strategies to learn generalizable features. which is a dominant line in this field. To name a few, some works learn domain-invariant feature representations through explicit feature alignment \cite{jin2020style,motiian2017unified,ghifary2015domain}, adversarial learning \cite{ganin2015unsupervised,gong2019dlow,li2018deep}, gradient-based methods \cite{li2018learning,balaji2018metareg,li2019feature}, causality-based methods \cite{DBLP:journals/corr/abs-2111-13420} or meta-learning based method \cite{wei2021toalign}, \etcno. In this work, we showcase a conceptually simple operation, \ieno, learnable filtration in the frequency domain, can significantly strengthen the generalization performance on unseen domains, verified on both the close-set classification and open-set retrieval tasks.

\subsection{Frequency Domain Learning}
Frequency analysis has been widely used in conventional digital image processing for decades \cite{pitas2000digital,baxes1994digital}. Recently, frequency-based operations, \egno, Fourier transform, set forth to be incorporated into deep learning methods for different purposes in four aspects: 1) accelerating the training or facilitating the optimization of DNNs \cite{mathieu2013fast,pratt2017fcnn,li2020falcon,prabhu2020butterfly,nair2020fast,chitsaz2020acceleration}; 2) achieving effective data augmentation \cite{yang2020fda,xu2021fourier,liu2021feddg,huang2021fsdr}; 3) learning informative representations of non-local receptive fields \cite{chi2020fast,rao2021global,yi2021contrastive,suvorov2022resolution,mao2021deep}; 4) helping analyze and understand some behaviors of DNNs \cite{yin2019fourier,xu2019training,xu2018understanding,wang2020high} as a tool. As introduced before, prior theoretical studies from the frequency perspective uncover that different frequency components are endowed with different priorities during training and contribute differently to the feature robustness. This inspires us to enhance the generalization ability of DNNs through modulating different frequency components.

In \cite{chi2020fast}, for intermediate features, a 1$\times$1 convolution in the frequency domain after FFT to learn global representations. However, such global representations capture global characteristics while losing local ones, thus have been demonstrated complementary with the features learned in the original latent space. To address this, a two-branch architecture is proposed in \cite{chi2020fast} to fuse these two kinds of features. This problem also exists in our work but is not our focus. Thereby, we adopt our proposed frequency filtering operation in the spectral transform branch of the two-branch architecture proposed in \cite{chi2020fast} for effectiveness evaluation. Besides, in \cite{rao2021global}, a learnable filter layer is adopted to self-attention (\ieno, transformer) to mix tokens representing different spatial locations, which may seem similar with ours at its first glance but is actually not. The learnable filter in \cite{rao2021global} is implemented by network parameters while that of ours is the network output thus instance-adaptive. We theoretically analyze and experimentally compare them in the following sections. Besides, with a different purpose from token mixing, we are devoted to improve the generalization ability of DNNs.

\subsection{Attention Mechanisms}
Attention has achieved great success in many visual tasks. It can be roughly categorized into selective attention \cite{hu2018squeeze,DBLP:conf/cvpr/WangWZLZH20,jaderberg2015spatial,bello2019attention,woo2018cbam,zhang2020relation,zhang2020multi,qin2021fcanet,zhu2022hmfca} and self-attention \cite{wang2018non,jaderberg2015spatial,carion2020end,DBLP:conf/iclr/DosovitskiyB0WZ21,arnab2021vivit,DBLP:conf/icml/BertasiusWT21,fan2021multiscale,patrick2021keeping} upon their working mechanisms. The former one explicitly learns a mask to enhance task-beneficial features and suppress task-unrelated features. In contrast, self-attention methods commonly take affinities of tokens as the attentions weights to refine the token representations via message passing, wherein the attention weights can be understood to model the importance of other tokens for the query token. Our proposed frequency filtering is implemented with a simple selective attention applied in the frequency domain for the intermediate features of DNNs. There have been a few primary attempts \cite{qin2021fcanet,zhu2022hmfca} exploiting frequency representations to learn more effective attention. In these works \cite{qin2021fcanet,zhu2022hmfca}, channel attention weights are proposed to learned from multiple frequency components of 2D DCT, where they are still used to modulate channels in the original feature space. In our work, we further investigate the frequency filtering where the the learning and using of attention weights are both in the frequency domain. We make the first endeavour to showcase the effectiveness of such a conceptually simple mechanism for the DG field, and would leave more delicate designs of attention model architectures for the frequency domain in our future work.

\section{Deep Frequency Filtering}


\subsection{Problem Definition and Core Idea}
In this paper, we aim to reveal that the generalization ability of DNNs to unseen domains can be significantly enhanced through an extremely simple mechanism, \ieno, an explicit frequency modulation in the latent space, named \ourfull~(\ours). To shed light on this core idea, we first introduce the problem definition of Domain Generalization (DG) as preliminaries. Given $K$ source domains $\mathcal{D}_{s}=\{D_s^1, D_s^2, \cdots, D_s^K\}$, where $D_s^k={(\mathbf{x}_i^k, y_i^k)}_{i=1}^{N_k}$ denotes the $k$-th domain 
consisting of $N_k$ samples $\mathbf{x}_i^k$ with their corresponding labels $y_i^k$, the goal of DG is to enable the model trained on source domains $\mathcal{D}_{s}$ perform as well as possible on unseen target domains $\mathcal{D}_{t}$, without additional model updating using the data in $\mathcal{D}_{t}$. When different domains share the same label space, it corresponds to a closed-set DG problem, otherwise an open-set problem.

As introduced before, the studies for the behaviors of DNNs from the frequency perspective \cite{xu2019training,xu2018understanding,wang2020high} have uncovered that the DNNs have different preferences for different frequency components of the learned intermediate features. The frequency characteristics affect the trade-off between robustness and accuracy \cite{wang2020high}. This inspires us to improve the generalization ability of DNNs through modulating different frequency components of different transfer difficulties across domains during training. Achieved by a simple filtering operation, transferable frequency components are enhanced while the components prejudice to cross-domain generalization are suppressed.

\begin{figure*}[t]

	\setlength{\abovecaptionskip}{0pt} 
\setlength{\belowcaptionskip}{-0pt}
	\begin{center}
		\includegraphics[width=0.92\linewidth]{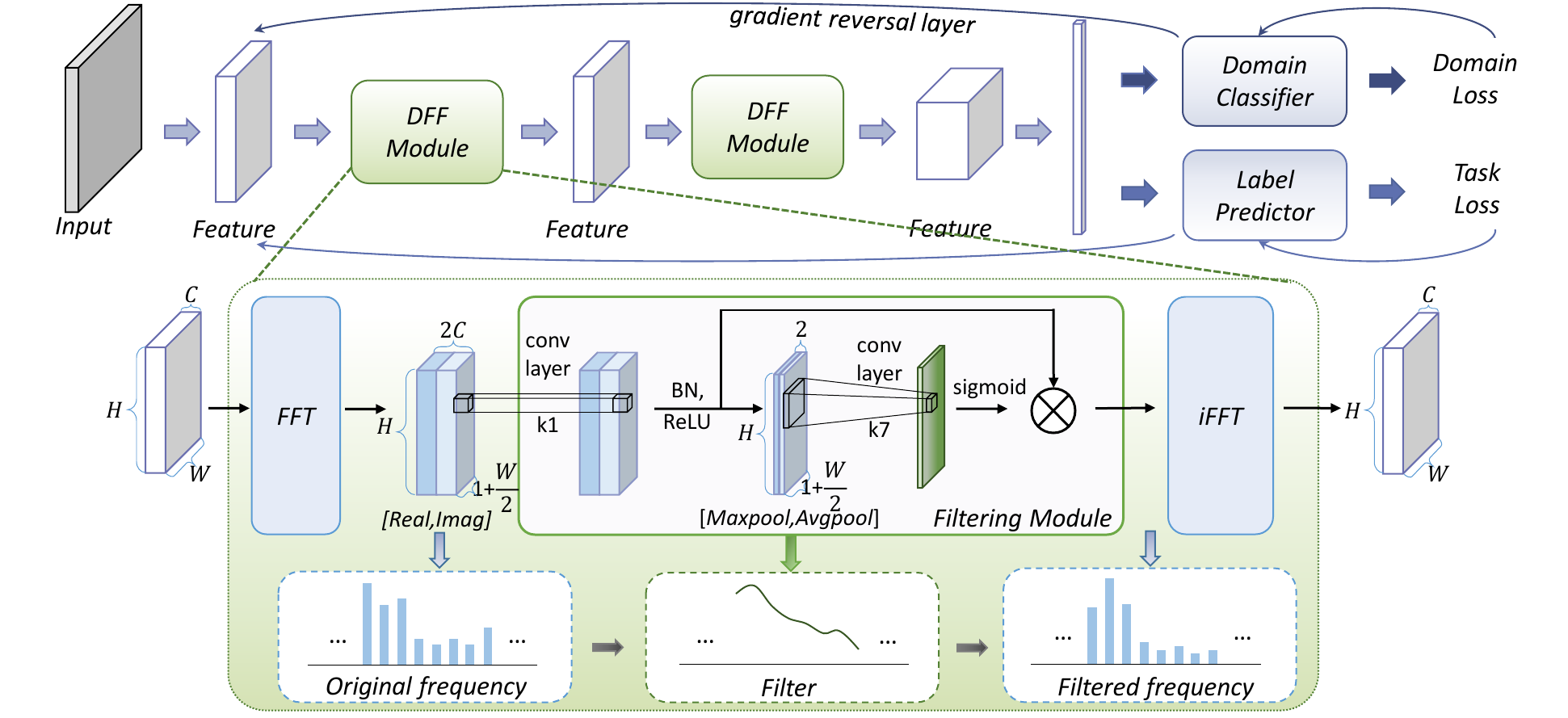}   
	\end{center}
	\caption{Illustration of our proposed Deep Frequency Filtering (DFF) module. DFF learns an instance-adaptive spatial mask to dynamically modulate different frequency components during training for learning generalizable features, which consists of three key operations: a 2D fast Fourier transform (FFT) to convert the input spatial features to the frequency domain, an filtering module to enhance the transferable components while suppressing the generalization-detrimental ones, and a 2D inverse FFT (iFFT) to map the features back to the orginal latent space. }
	\label{fig:framework}
\end{figure*}



\subsection{Latent Frequency Representations}
\label{sec:LFR}

Different from previous frequency-based methods \cite{yang2020fda,liu2021feddg,xu2021fourier} applied in the pixel space (\ieno, the side of inputs), we adopt our proposed filtering operation in the latent space. In this section, we briefly recall a conventional signal processing tool Fast Fourier Transform (FFT). We adopt it for obtaining the feature representations in the frequency domain, then discuss the characteristics of such representations.  

Given the intermediate features $\mathbf{X}\in \mathbb{R}^{C \times H \times W}$, we perform a 2D fast Fourier transform (\ieno, an accelerated version \cite{cooley1965algorithm} of 2D discrete Fourier transform) for each channel independently to get the corresponding frequency representations $\mathbf{X}_{F} \in \mathbb{R}^{2C \times H \times (\lfloor \frac{W}{2} \rfloor + 1)}$. We formulate this transform $\mathbf{X}_{F}$=$FFT(\mathbf{X})$ as below (where the channel dimension is omitted for brevity):
\begin{equation} 
\label{equ:1}
\mathbf{X}_{F}(x, y) = \sum_{h=0}^{H-1} \sum_{w=0}^{W-1}\mathbf{X}(h, w) e^{-j2\pi\left(x \frac{h}{H}+y \frac{w}{W}\right)}.
\end{equation}
The frequency representation $\mathbf{X}_{F}$ can be transferred to the original feature space via an inverse FFT, succinctly expressed as $\mathbf{X}$=$iFFT(\mathbf{X}_{F})$, which can be formulated as: 
\begin{equation} 
\label{equ:2}
\mathbf{X}(h, w) = \frac{1}{H\cdot W} \sum_{h=0}^{H-1} \sum_{w=0}^{W-1} \mathbf{X}_{F}(x, y) e^{j2\pi\left(x \frac{h}{H}+y \frac{w}{W}\right)}.
\end{equation}
The $\mathbf{X}_{F} \in \mathbb{R}^{2C \times H \times (\lfloor \frac{W}{2} \rfloor + 1)}$ above denotes the frequency representation of $\mathbf{X}\in \mathbb{R}^{C \times H \times W}$, which concatenates the real and imaginary parts after FFT (each one has $C$ channels). Besides, thanks to the conjugate symmetric property of FFT, $\mathbf{X}_{F}$ only needs retain the half of spatial dimensions thus has a spatial resolution of $H \times (\lfloor \frac{W}{2} \rfloor + 1)$. For the frequency representation $\mathbf{X}_{F}$, there are two utilizable properties: 1) Different frequency components of the original feature $\mathbf{X}$ are decomposed into elements at different spatial locations of $\mathbf{X}_{F}$, which could be viewed as a frequency-based disentanglement and re-arrangement for $\mathbf{X}$. This property makes the learning in frequency domain efficient in practice, and more importantly, allows us to achieve frequency filtering with a simple devised spatial attention module. 2) $\mathbf{X}_{F}$ is a naturally global feature representation, as discussed in \cite{chi2020fast}, which can facilitate the suppression of globally distributed domain-specific information, such as illumination, imaging noises, \etcno. Next, we shed light on the specific filtering operation on $\mathbf{X}_{F}$.

\subsection{Latent-space Frequency Filtering}
\label{sec:FFF}

Our goal is to adaptively modulate different frequency components over different network depths during training. We thus propose to apply a frequency filtering operation on $\mathbf{X}_{F}$ to enhance the transferable components while suppressing the generalization-detrimental ones. Thanks to the first hallmark of $\mathbf{X}_{F}$ discussed in Sec. \ref{sec:LFR}, the frequency filtering operation is allowed to be implemented with a spatial attention on $\mathbf{X}_{F}$. Given a frequency representation $\mathbf{X}_F \in \mathbb{R}^{2C \times H \times (\left\lfloor\frac{W}{2}\right\rfloor+1)}$, the proposed frequency filtering mechanism is formulated as follows:
\begin{equation} \label{equ:3}
\mathbf{X}'_F =\mathbf{X}_F \otimes \mathbf{M}_S(\mathbf{X}_F),
\end{equation}
where $\otimes$ denotes element-wise multiplication. $\mathbf{M}_S(\cdot)$ refers to the attention module to learn a spatial mask with a resolution of $H \times (\left\lfloor\frac{W}{2}\right\rfloor+1)$. This mask is copied along the channel dimension of $\mathbf{X}_F$ accordingly for the element-wise multiplication, filtering out the components adverse to the generalization in $\mathbf{X}_F$. The frequency feature after filtering is denoted by $\mathbf{X}'_F$. Our contributions lie in revealing such a frequency filtering operation in the latent space can bring impressive improvements for DG, although using a lightweight attention architecture designed for the features in original latent space \cite{woo2018cbam} to implement $\mathbf{M}_S(\cdot)$. This provides another alternative to the field of DG, which is conceptually simple, significantly effective, but previously unexplored. Besides, we further conduct an empirical study to investigate the effectiveness of different attention types in implementing our conceptualized deep frequency filtering in the experiment section. The specific architecture design for the attention module is not our current focus, but is worth being explored in the future. 

Here, we introduce an extremely simple instantiation for Eq. (\ref{equ:3}). We use this for verifying our proposed concept \emph{deep frequency filtering} in this paper. For $\mathbf{X}_F \in \mathbb{R}^{2C \times H \times (\left\lfloor\frac{W}{2}\right\rfloor+1)}$ that consists of real and imaginary parts after FFT, inspired by the attention architecture design in \cite{woo2018cbam}, we first adopt a $1 \times 1$ convolutional layer followed by Batch Normalization (BN) and ReLU activation to project $\mathbf{X}_F$ to an embedding space for the subsequent filtration. After embedding, as shown in Fig.~\ref{fig:framework}, we follow the spatial attention architecture design in \cite{woo2018cbam} to aggregate the information of $\mathbf{X}_F$ over channels using both average-pooling and max-pooling operations along the channel axis, generating two frequency descriptors denoted by $\mathbf{X}_F^{avg}$ and $\mathbf{X}_F^{max}$, respectively. These two descriptors can be viewed as two compact representations of $\mathbf{X}_F$ in which the information of each frequency component is compressed separately by the pooling operations while the spatial discriminability is still preserved. We then concatenate $\mathbf{X}_F^{avg}$ with $\mathbf{X}_F^{max}$ and use a large-kernel $7 \times 7$ convolution layer followed by a sigmoid function to learn the spatial mask. Mathematically, this instantiation can be formulated as:
\begin{equation} \label{equ:4}
\mathbf{X}'_F =\mathbf{X}_F \otimes \sigma(\operatorname{Conv}_{7\times 7}([\operatorname{AvgPool}(\mathbf{X}_F),\operatorname{MaxPool}(\mathbf{X}_F)])), \\
\end{equation}
where $\sigma$ denotes the sigmoid function. The $[\cdot,\cdot]$ is a concatenation operation. $\operatorname{AvgPool}(\cdot)$ and $\operatorname{MaxPool}(\cdot)$ denote the average and max pooling operations, respectively. $\operatorname{Conv}_{7\times7}(\cdot)$ is a convolution layer with the kernel size of $7$. Albeit using a large-size kernel, the feature $[\operatorname{AvgPool}(\mathbf{X}_F),\operatorname{MaxPool}(\mathbf{X}_F)]$ has only two channels through the information squeeze by pooling operations such that this step is still very computationally efficient in practice. We omit the embedding of $\mathbf{X}_F$ in this formulation for brevity. We believe using more complex attention architectures, such as \cite{zhang2020relation,misra2021rotate,dai2021coatnet}, is of the potentials to achieve higher improvements, and we expect more effective instantiations of our conceptualized Deep Frequency Filtering.

\paragraph{Discussion.}
The proposed Deep Frequency Filtering is conceptually new design to achieve instance-adaptive frequency modulation in the latent space of DNNs. It also corresponds to a novel neural operation albeit using an off-the-shelf architecture design as an exampled instantiation. Compared to prior frequency-domain works \cite{chi2020fast,rao2021global}, we make the first endeavour to introduce an explicit instance-adaptive frequency selection mechanism into the optimization of DNNs. From the perspective of attention, conventional attention designs \cite{woo2018cbam,zhang2020relation,hu2018squeeze,wang2018non} learn masks from deep features in the original latent space, and adopt the learned masks to these features themselves to achieve feature modulation. FcaNet \cite{qin2021fcanet} strives to a further step by learning channel attention weights from the results of frequency transform. But the learned attention weights are still used for the original features. In this aspect, we are the first to learn attention weights from frequency representations and also use the learned masks in the frequency domain to achieve our conceptualized frequency filtering. 

\subsection{Post-filtering Feature Restitution}


The features captured in the frequency domain have been demonstrated to be global and complementary to the local ones captured in the original latent space in \cite{chi2020fast}. Thus, a simple two-branch is designed to exploit this complementarity to achieve an ensemble of both local and global features in \cite{chi2020fast}. This architecture is naturally applicable to the restitution of complementary local features as a post-filtering refinement in the context of our proposed concept. We thus evaluate the effectiveness of our proposed method on top of the two-branch architecture in \cite{chi2020fast}. Specifically, similar to \cite{chi2020fast}, we split the given intermediate feature $\mathbf{X} \in \mathbb{R}^{C \times H \times W}$ along its channel dimension into $\mathbf{X}^{g} \in \mathbb{R}^{rC \times H \times W}$ and $\mathbf{X}^{l} \in \mathbb{R}^{(1-r)C \times H \times W}$. The two-branch architecture can be formulated as:
\begin{equation} \label{equ:5}
    Y^l = f_l(\mathbf{X}^{l}) + f_{g\rightarrow l}(\mathbf{X}^{g}),\quad
    Y^g = f_g(\mathbf{X}^{g}) + f_{l\rightarrow g}(\mathbf{X}^{l}),
\end{equation}
where $f_l(\cdot)$, $f_g(\cdot)$, $f_{l\rightarrow g}(\cdot)$ and $f_{g\rightarrow l}(\cdot)$ denote four different transformation functions. Among them, $f_l(\cdot)$, $f_{l\rightarrow g}(\cdot)$ and $f_{g\rightarrow l}(\cdot)$ are three regular convolution layers. In \cite{chi2020fast}, the $f_g(\cdot)$ corresponds to the spectral transform implemented by their proposed convolution operation in the frequency domain. On top of it, we evaluate the effectiveness of our proposed \ours~by adding this operation into the spectral transform branch of this architecture to achieve an explicit filtering operation in the frequency domain. The contribution on this two-branch architecture design belongs to \cite{chi2020fast}. 

\subsection{Model Training} 
In addition to commonly used task-related loss functions (for classification or retrieval), we train a domain classifier with a domain classification loss and adopt a gradient reversal layer \cite{ganin2015unsupervised}. These are commonly used in DG research works for explicitly encouraging the learning of domain-invariant features and the suppression for features conducive to domain generalization. The feature extractor is optimized for minimizing the task losses while maximizing the domain classification loss simultaneously.

\section{Experiments}

\subsection{Datasets and Settings}
\label{section:dataset}
We evaluate the effectiveness of our proposed Deep Frequency Filtering (DFF) for Domain Generalization (DG) on \textbf{Task-1}: the close-set classification task and \textbf{Task-2}: the open-set retrieval task, \ieno, person re-identification (ReID). 



For Task-1, Office-Home dataset~\cite{li2017deeper} is a commonly used domain generalization (DG) benchmark on the task of classification. It consists of 4 domains (\ieno, Art (Ar), Clip Art (Cl), Product (Pr), Real-World (Rw)). Among them, three domains are used for training and the remaining one is considered as the unknown target domain for testing. 

For Task-2, person ReID is a representative open-set retrieval task, where different domains do not share their label space. 
i) following \cite{zhao2020learning,liu2022debiased}, we take four large-scale datasets (CUHK-SYSU (CS) \cite{xiao2017joint}, MSMT17 (MS) \cite{wei2018person}, CUHK03 (C3) \cite{li2014deepreid} and Market-1501 (MA) \cite{zheng2015scalable}). For evaluation, a model is trained on three domains and tested on the remaining one.
ii) several large-scale ReID datasets \egno, CUHK02 \cite{li2013locally}, CUHK03, Market-1501 and CUHK-SYSU, are viewed as multiple source domains. 
Each small-scale ReID dataset including VIPeR \cite{gray2008viewpoint}, PRID \cite{hirzer2011person}, GRID \cite{loy2009multi} and iLIDS \cite{zheng2009associating} is used as an unseen target domain. 
To comply with the General Ethical Conduct, we exclude DukeMTMC from the source domains. 

We adopt ResNet-18 and ResNet-50 \cite{he2016deep} as our
backbone for Task-1 and Task-2, respectively. All reported results are obtained by the averages of five runs. We provide more implementation details in the supplementary material.

\subsection{Ablation Study}


\begin{table}[t]
  \caption{Performance comparisons of our proposed \ourfull~(\ours) with the baselines and the models with deep filtering in the original feature space. ``\emph{Base}'' refers to the vanilla ResNet baseline. In ``\emph{SBase}'', we use ResNet-based FFC in \cite{chi2020fast}, serving as a strong baseline. ``\emph{Ori-F}'' refers to a filtering operation in the original feature space, adopted in the local branch $f_l$ and global branch $f_g$, respectively. When adopted in $f_g$, the FFT/iFFT operations are discarded from $f_g$ so that the filtering is in the original feature space instead of the frequency space. ``\emph{Fre-F}'' represents our proposed \ourfull. }
  \label{table:ablationDFF}
  \centering
  \scalebox{0.73}{\begin{tabular}{c|cccccc}
\hline
                                                                                                                             
                      \multirow{3}{*}{Method}   & \multicolumn{6}{c}{Source$\rightarrow$Target}                                                                                                                                                          \\  \cline{2-7} 
                        &    \multicolumn{2}{c|}{MS+CS+C3$\rightarrow$MA}                             & \multicolumn{2}{c|}{MS+MA+CS$\rightarrow$C3}                             & \multicolumn{2}{c}{MA+CS+C3$\rightarrow$MS}         \\ \cline{2-7} 
                        &\quad mAP           & \multicolumn{1}{c|}{R1}            & \quad mAP           & \multicolumn{1}{c|}{R1}            & \quad mAP           & R1            \\ \hline
 \multicolumn{1}{l|}{Base}                &  \quad 59.4          & \multicolumn{1}{c|}{83.1}         & \quad   30.3            & \multicolumn{1}{c|}{29.1}              & \quad 18.0              &   \multicolumn{1}{c}{41.9}     \\
 \multicolumn{1}{l|}{SBase (FFC) }                 &     \quad 66.2          & \multicolumn{1}{c|}{84.7}          & \quad   35.8            & \multicolumn{1}{c|}{35.4}              & \quad  19.4             &    44.8       \\    
 \hline
 \multicolumn{1}{l|}{Ori-F in $f_l$}                      &  \quad 66.9          & \multicolumn{1}{c|}{85.0}          & \quad   36.2            & \multicolumn{1}{c|}{35.9}              & \quad  19.8             &    45.1           \\ 
  \multicolumn{1}{l|}{Ori-F in $f_g$}                      &  \quad 61.9          & \multicolumn{1}{c|}{83.5}          & \quad   32.7            & \multicolumn{1}{c|}{31.9}              & \quad  18.4             &    42.8           \\  \hline
\rowcolor{gray!10}  \multicolumn{1}{l|}{Fre-F (Ours)}               & \quad \textbf{71.1}         & \multicolumn{1}{c|}{\textbf{87.1}}         & \quad     \textbf{41.3}         & \multicolumn{1}{c|}{\textbf{41.1}}              & \quad        \textbf{25.1}       &   \multicolumn{1}{c}{\textbf{50.5}}     \\ \hline
\end{tabular}
}

  \vspace{-0.1cm}
\end{table}

\begin{table}[t]

	\begin{center}

  \caption{Performance comparisons of different implementations of the frequency filtering operation. ``\emph{Task.}'' refers to the filtering operation using a task-level attention mask where the mask is implemented with network parameters and is shared over different instances. ``\emph{Ins.}'' denotes the filtering operation using learned instance-adaptive masks. ``\emph{C}'' and ``\emph{S}'' represents the filtering performed along the channel and spatial dimensions, respectively.}

  \label{table:TASK}
  \centering
  \scalebox{0.73}{\begin{tabular}{c|cccccc}
\hline
 
                      \multirow{3}{*}{Method}   & \multicolumn{6}{c}{Source$\rightarrow$Target}                                                                                                                                                          \\  \cline{2-7} 
                        &    \multicolumn{2}{c|}{MS+CS+C3$\rightarrow$MA}                             & \multicolumn{2}{c|}{MS+MA+CS$\rightarrow$C3}                             & \multicolumn{2}{c}{MA+CS+C3$\rightarrow$MS}         \\ \cline{2-7} 
                        &\quad mAP           & \multicolumn{1}{c|}{R1}            & \quad mAP           & \multicolumn{1}{c|}{R1}            & \quad mAP           & R1            \\ \hline
 \multicolumn{1}{l|}{Base}                &  \quad 59.4          & \multicolumn{1}{c|}{83.1}         & \quad   30.3            & \multicolumn{1}{c|}{29.1}              & \quad 18.0              &   \multicolumn{1}{c}{41.9}     \\ \hline
 \multicolumn{1}{l|}{Task.(C) }                 &     \quad 62.7          & \multicolumn{1}{c|}{80.0}          & \quad   32.1           & \multicolumn{1}{c|}{31.4}              & \quad  19.5             &    44.9     \\    

 \multicolumn{1}{l|}{Task.(S)}             & \quad  68.6     & \multicolumn{1}{c|}{85.8 }         & \quad    37.0             & \multicolumn{1}{c|}{36.3 }              & \quad     20.8         &   \multicolumn{1}{c}{ 45.4}     \\   
 \multicolumn{1}{l|}{Ins.(C)}                     &  \quad 69.8          & \multicolumn{1}{c|}{86.2}          & \quad   36.4           & \multicolumn{1}{c|}{35.9}              & \quad  21.0             &    45.7          \\  \hline
\rowcolor{gray!10}  \multicolumn{1}{l|}{Ins.(S) (Ours)}               & \quad \textbf{71.1}         & \multicolumn{1}{c|}{\textbf{87.1}}         & \quad     \textbf{41.3}         & \multicolumn{1}{c|}{\textbf{41.1}}              & \quad        \textbf{25.1}       &   \multicolumn{1}{c}{\textbf{50.5}}     \\ \hline
\end{tabular}
  }

		\end{center}
	\vspace{-0.50cm}
\end{table}%

\subsubsection{The effectiveness of \ours}
To investigate the effectiveness of our proposed \ourfull~(\ours), we compare it with the ResNet baselines (\emph{Base}) and the ResNet-based FFC \cite{chi2020fast} models (\emph{SBase}) that serve as the strong baselines, respectively. The experiment results are in Table \ref{table:ablationDFF}. 
There are two crucial observations: 1) Our \emph{Fre-F (\ours)} consistently outperforms \emph{Base} by a clear margin. 
This demonstrates the effectiveness of our proposed method. 
With a simple instantiation, our proposed \ourfull~operation can significantly 
improve the generalization ability of DNNs. 
2) Our \emph{Fre-F (\ours)} brings more improvements than \emph{SBase (FFC)}. 
This indicates that frequency-domain convolutions are inadequate for model generalization. 
Instead, \ours~is an effective solution for classification generalization task. 


\subsubsection{Frequency-domain v.s. Original-domain filtering}

Feature filtering operations can be implemented in either the original feature domain or the frequency domain.
We conduct experiments to study the impact of different implementation in both domains. 
Table~\ref{table:ablationDFF} indicates that the proposed frequency-domain filtering (\emph{Fre-F}) outperforms the original-domain feature filtering (\emph{Ori-F}).  
This demonstrates the importance of modulating different frequency components in the latent space for domain generalization.

\subsubsection{The importance of instance-adaptive attention}
Our proposed \ours~is allowed to be implemented using a spatial attention on the frequency representations of features. In Table~\ref{table:TASK}, we compare the performance of adopting task-level and instance-level attention mask as well as channel and spatial filtering, respectively. 
We can observe that \emph{Ins. (S)} clearly and consistently outperforms \emph{Task. (S)} over all settings on both tasks. 
The results suggest that using instance-adaptive attention mask is essential for implementing our proposed~\ours. 
We observe that different instances correspond to diversified frequency components in the feature space. 
Thus, a task-level attention mask is weak for all instances, which may explain the performance gaps in Table~\ref{table:TASK}. 
We need to perform the modulation of feature frequency components with instance-adaptive weights.

\subsubsection{Spatial v.s. Channel}
\ours~can be implemented with a spatial attention module, since different spatial positions of the frequency representations correspond to different frequencies. 
Admittedly, we can also adopt the channel attention to the frequency representations, which can be viewed as a representation refinement within each frequency component rather than perform \ourfull. 
The results in Table \ref{table:TASK} show that spatial attention consistently outperforms channel attention in both task-level and instance-level settings. 
This indicates that frequency filtering (or selection) is more important than the refinement of frequency-domain representations for domain generalization. 

\begin{table}[t]
	\begin{center}

  \caption{The ablation results on the influence of the two-branch architecture. ``\emph{FFC}'' refers to the frequency-domain convolution work \cite{chi2020fast} without our proposed filtering operation. ``\emph{only $f_g$}'' denotes the setting in which we adopt the \emph{FFC} or our proposed \emph{\ours} over the complete feature without the splitting along the channel dimension, corresponding to using a single branch architecture. ``$f_l$ + $f_g$'' represents the setting in which we split the feature along the channel dimension and adopt frequency-domain operations (\emph{FFC} or our \emph{\ours}) on one half of them.}

  \label{table:twobranch}
  \centering
  \scalebox{0.73}{\begin{tabular}{c|cccccc}
\hline
                                                                                                                             
                      \multirow{3}{*}{Method}   & \multicolumn{6}{c}{Source$\rightarrow$Target}                                                                                                                                                          \\  \cline{2-7} 
                        &    \multicolumn{2}{c|}{MS+CS+C3$\rightarrow$MA}                             & \multicolumn{2}{c|}{MS+MA+CS$\rightarrow$C3}                             & \multicolumn{2}{c}{MA+CS+C3$\rightarrow$MS}         \\ \cline{2-7} 
                        &\quad mAP           & \multicolumn{1}{c|}{R1}            & \quad mAP           & \multicolumn{1}{c|}{R1}            & \quad mAP           & R1            \\ \hline
 \multicolumn{1}{l|}{Base}                &  \quad 59.4          & \multicolumn{1}{c|}{83.1}         & \quad   30.3            & \multicolumn{1}{c|}{29.1}              & \quad 18.0              &   \multicolumn{1}{c}{41.9}     \\ \hline
 \multicolumn{1}{l|}{FFC(only $f_g$)}                 & \quad   60.1            & \multicolumn{1}{c|}{80.4}              & \quad    26.1           & \multicolumn{1}{c|}{24.8}              & \quad    17.3           &     40.2     \\    

 \multicolumn{1}{l|}{FFC($f_l+f_g$)}                       &     \quad 66.2          & \multicolumn{1}{c|}{84.7}          & \quad   35.8            & \multicolumn{1}{c|}{35.4}              & \quad  19.4             &    44.8       \\      
 \multicolumn{1}{l|}{Ours(only $f_g$)}  &                   \quad  64.2             & \multicolumn{1}{c|}{83.4}              & \quad    29.3           & \multicolumn{1}{c|}{28.1}              & \quad    17.6           &      40.4      \\  \hline
\rowcolor{gray!10}  \multicolumn{1}{l|}{Ours($f_l+f_g$)}               & \quad \textbf{71.1}         & \multicolumn{1}{c|}{\textbf{87.1}}         & \quad     \textbf{41.3}         & \multicolumn{1}{c|}{\textbf{41.1}}              & \quad        \textbf{25.1}       &   \multicolumn{1}{c}{\textbf{50.5}}     \\ \hline
\end{tabular}
  }

		\end{center}
	\vspace{-0.50cm}
\end{table}%

\begin{table*}[h]
\vspace{-0.5em}
\begin{center}
\caption{Performance (classification accuracy \%) comparison with the state-of-the-art methods on close-set classification task. We use ResNet-18 as backbone. Best in bold.}
\label{table:OfficeHome}
\vspace{-0.8em}
\scalebox{0.73}{
\begin{tabular}{l|c|c|c|c|c}
\hline
\multirow{2}{*}{Method} &\multicolumn{4}{c|}{Source$\rightarrow$Target}& \multirow{2}{*}{Avg} \\ \cline{2-5}& {Cl,Pr,Rw$\rightarrow$Ar} & {Ar,Pr,Rw$\rightarrow$Cl} & {Ar,Cl,Rw$\rightarrow$P} & Ar,Cl,Pr$\rightarrow$Rw &\\ \hline
Baseline & 52.2& 45.9& 70.9& 73.2& 60.5\\
CCSA \cite{motiian2017unified}& 59.9& 49.9& 74.1& 75.7& 64.9\\
D-SAM \cite{d2018domain}& 58.0& 44.4& 69.2& 71.5& 60.8\\
MMD-AAE \cite{li2018domain}& 56.5& 47.3& 72.1& 74.8& 62.7\\
CrossGrad \cite{shankar2018generalizing}& 58.4& 49.4& 73.9& 75.8& 64.4\\
JiGen \cite{carlucci2019domain}& 53.0& 47.5& 71.5& 72.8& 61.2\\
RSC \cite{huang2020self}& 58.4& 47.9& 71.6& 74.5& 63.1\\
MixStyle \cite{zhou2021domain}& 58.7& 53.4& 74.2& 75.9& 65.5\\
\rowcolor{gray!10} Ours& \textbf{65.4}& \textbf{53.7}& \textbf{74.9}& \textbf{76.5 }& \textbf{67.6}\\ \hline
\end{tabular}}
\vspace{-0.5em}
\end{center}
\end{table*}

\begin{table*}[h]
\vspace{-0.5em}
  \caption{Performance (\%) comparison with the state-of-the-art methods on the open-set person ReID task. “Source” refers to the multiple training datasets,
i.e., Market-1501 (MA), DukeMTMC-reID (D), CUHK-SYSU (CS), CUHK03 (C3) and CUHK02 (C2). “All” represents using MA+D+CS+C3+C2 as source domains. We do not include DukeMTMC-reID (D) in the training domains since this dataset has been discredited by the creators, denoted as “All w/o D”.}
  \label{table:small}
  \centering
  \vspace{-0.8em}
 \scalebox{0.71}{\begin{tabular}{c|c|cc|cc|cc|cc|cc}
\hline
\multirow{2}{*}{Method}  & \multirow{2}{*}{Source} & \multicolumn{2}{c|}{Target:VIPeR(V)}          & \multicolumn{2}{c|}{Target:PRID(P)}           & \multicolumn{2}{c|}{Target:GRID(G)}           & \multicolumn{2}{c|}{Target:iLIDS(I)}          & \multicolumn{2}{c}{Mean of V,P,G,I}           \\ \cline{3-12} 
                        &                                            & R1                     & mAP           & R1                      & mAP           & R1                    & mAP           & R1                    & mAP           & R1                     & mAP           \\ \hline
\multicolumn{1}{l|}{CDEL     \cite{lin2021domain}  }                                   &  All            & 38.5                  & -             & 57.6                  & -             & 33.0                & -             & 62.3                   & -             & 47.9                 & -             \\
\multicolumn{1}{l|}{DIMN   \cite{song2019generalizable}  }                                     &  All            & 51.2               & 60.1          & 39.2               & 52.0          & 29.3             & 41.1          & 70.2                & 78.4          & 47.5               & 57.9          \\
\multicolumn{1}{l|}{DDAN  \cite{chen2021dual} }                                       &  All            & 56.5                & 60.8          & 62.9               & 67.5          & 46.2            & 50.9          & 78.0             & 81.2          & 60.9               & 65.1          \\
\multicolumn{1}{l|}{RaMoE   \cite{dai2021generalizable}  }                                    &  All            & 56.6                 & 64.6          & 57.7                  & 67.3          & 46.8                & 54.2          & 85.0                    & 90.2          & 61.5                 & 69.1          \\
\multicolumn{1}{l|}{SNR    \cite{jin2020style} }                                      &  All            & 49.2            & 58.0          & 47.3              & 60.4          & 39.4              & 49.0          & 77.3             & 84.0          & 53.3            & 62.9          \\
\multicolumn{1}{l|}{CBN    \cite{zhuang2020rethinking}  }                                     &  All            & 49.0                & 59.2          & 61.3              & 65.7          & 43.3              & 47.8          & 75.3                 & 79.4          & 57.2                & 63.0          \\
\multicolumn{1}{l|}{Person30K   \cite{bai2021person30k}}                                  &  All            & 53.9               & 60.4          & 60.6                 & 68.4          & 50.9               & 56.6          & 79.3                & 83.9          & 61.1              & 67.3          \\
\multicolumn{1}{l|}{DIR-ReID   \cite{zhang2021learning}  }                                 &  All            & 58.3              & 62.9          & 71.1            & 75.6          & 47.8               & 52.1          & 74.4              & 78.6          & 62.9              & 67.3          \\
\multicolumn{1}{l|}{MetaBIN    \cite{choi2021meta}   }                                &  All            & 56.2                & 66.0          & 72.5               & 79.8          & 49.7                  & 58.1          & 79.7                & 85.5          & 64.5                & 72.4          \\ \hline
\multicolumn{1}{l|}{$\mathrm{QAConv}_{50}$  \cite{liao2020interpretable} }                                    & All w/o D              & 57.0                  & 66.3          & 52.3                  & 62.2          & 48.6            & 57.4          & 75.0               & 81.9          & 58.2              & 67.0          \\
\multicolumn{1}{l|}{$\mathrm{M}^{3}\mathrm{L} $    \cite{zhao2021learning} }                                     & All w/o D            & 60.8               & 68.2          & 55.0             & 65.3          & 40.0                & 50.5          & 65.0             & 74.3          & 55.2                & 64.6          \\
\multicolumn{1}{l|}{MetaBIN   \cite{choi2021meta}       }                               & All w/o D              & 55.9                  & 64.3          & 61.2                  & 70.8          & 50.2            & 57.9          & 74.7              & 82.7          & 60.5               & 68.9          \\
\rowcolor{gray!10}\multicolumn{1}{l|}{Ours      }                                    & All w/o D              & \textbf{65.7} & \textbf{74.2} & \textbf{71.8}  & \textbf{78.6} & \textbf{56.4} & \textbf{65.5} & \textbf{83.6} & \textbf{88.3} & \textbf{69.4} & \textbf{76.7} \\ \hline
\end{tabular}
  }
  \vspace{-0.5em}
\end{table*}

\begin{table*}[t]
  \vspace{-0.5em}
  \caption{Performance (\%) comparison with the state-of-the-art methods on the open-set person ReID task. Evalustion on four large-scale person ReID benchmarks including Market-1501 (MA), Cuhk-SYSC (CS), CUHK03 (C3) and MSMT17 (MS). `Com-'' refers to combining the train and test sets of source domains for training. The $\mathrm{M}^{3}\mathrm{L}$ with a superscript $^*$ denote the model adopting IBN-Net50 as backbone. Without this superscript, ResNet-50 is taken as the backbone.  }
  \label{table:big}
  \centering
    \vspace{-0.8em}
  \scalebox{0.76}{
  \begin{tabular}{c|c|cc|c|cc|c|cc}
\hline
\multirow{2}{*}{Method} & \multirow{2}{*}{Source}                                                    & \multicolumn{2}{c|}{Market-1501} & \multirow{2}{*}{Source}                                                   & \multicolumn{2}{c|}{CUHK03}   & \multirow{2}{*}{Source}                                                   & \multicolumn{2}{c}{MSMT17}    \\ \cline{3-4} \cline{6-7} \cline{9-10}  
                        &                                                                            & mAP             & R1             &                                                                           & mAP           & R1            &                                                                           & mAP           & R1            \\ \hline
\multicolumn{1}{l|}{SNR   \cite{jin2020style} }                   & \multirow{6}{*}{MS+CS+C3}                                                  & 34.6            & 62.7           & \multirow{6}{*}{MS+CS+MA}                                                  & 8.9           & 8.9           & \multirow{6}{*}{CS+MA+C3}                                                  & 6.8           & 19.9          \\
\multicolumn{1}{l|}{$\mathrm{M}^{3}\mathrm{L} $ \cite{zhao2021learning}}                     &                                                                            & 58.4            & 79.9           &                                                                           & 20.9          & 31.9          &                                                                           & 15.9          & 36.9          \\
\multicolumn{1}{l|}{$\mathrm{M}^{3}\mathrm{L} $*  \cite{zhao2021learning}  }                 &                                                                            & 61.5            & 82.3           &                                                                           & 34.2          & 34.4          &                                                                           & 16.7          & 37.5          \\
\multicolumn{1}{l|}{$\mathrm{QAConv}_{50}$ \cite{liao2020interpretable}  }               &                                                                            & 63.1            & 83.7           &                                                                           & 25.4          & 24.8          &                                                                           & 16.4          & 45.3          \\
\multicolumn{1}{l|}{MetaBIN   \cite{choi2021meta}  }              &                                                                            & 57.9            & 80.0           &                                                                           & 28.8          & 28.1          &                                                                           & 17.8          & 40.2          \\
\rowcolor{gray!10}\multicolumn{1}{l|}{ Ours  }                  &                                                                            & \textbf{71.1}   & \textbf{87.1}  &                                                                           & \textbf{41.3} & \textbf{41.1} &                                                                           & \textbf{25.1} & \textbf{50.5} \\ \hline
\multicolumn{1}{l|}{SNR      \cite{jin2020style}    }             & \multirow{6}{*}{\begin{tabular}[c]{@{}c@{}}Com-\\ (MS+CS+C3)\end{tabular}} & 52.4            & 77.8           & \multirow{6}{*}{\begin{tabular}[c]{@{}c@{}}Com-\\ (MS+CS+MA)\end{tabular}} & 17.5          & 17.1          & \multirow{6}{*}{\begin{tabular}[c]{@{}c@{}}Com-\\ (CS+MA+C3)\end{tabular}} & 7.7           & 22.0          \\
\multicolumn{1}{l|}{$\mathrm{M}^{3}\mathrm{L} $  \cite{zhao2021learning}   }                  &                                                                            & 61.2            & 81.2           &                                                                           & 32.3          & 33.8          &                                                                           & 16.2          & 36.9          \\
\multicolumn{1}{l|}{$\mathrm{M}^{3}\mathrm{L} $* \cite{zhao2021learning}    }                 &                                                                            & 62.4            & 82.7           &                                                                           & 35.7          & 36.5          &                                                                           & 17.4          & 38.6          \\
\multicolumn{1}{l|}{$\mathrm{QAConv}_{50}$    \cite{liao2020interpretable}  }            &                                                                            & 66.5            & 85.0           &                                                                           & 32.9          & 33.3          &                                                                           & 17.6          & 46.6          \\
\multicolumn{1}{l|}{MetaBIN  \cite{choi2021meta}    }           &                                                                            & 67.2            & 84.5           &                                                                           & 43.0          & 43.1          &                                                                           & 18.8          & 41.2          \\
 \rowcolor{gray!10} \multicolumn{1}{l|}{Ours  }                  &                                                                            & \textbf{81.0}   & \textbf{92.3}  &                                                                           & \textbf{51.5} & \textbf{51.2} &                                                                           & \textbf{25.3} & \textbf{51.8} \\ \hline
\end{tabular}}

    \vspace{-0.5em}
\end{table*}

\subsubsection{The Influence of the Two-branch Architecture}
As mentioned above, we adopt the two-branch architecture proposed by FFC \cite{chi2019fast} as a base architecture and apply our frequency filtering mechanism to the spectral transform branch. 
As shown in Table \ref{table:twobranch}, the performances of \emph{FFC(only $f_g$)} and \emph{FFC($f_l+f_g$)} are inferior to \emph{Ours(only $f_g$)} and \emph{Ours($f_l+f_g$)} respectively, which indicates that simply filtering frequency components of features can bring striking improvement of the generalization ability. 
And the performance gap between \emph{Ours($f_l+f_g$)} and \emph \emph{Ours(only $f_g$)} demonstrates that the two-branch structure for Post-filtering Feature Restitution can restore the globally filtered features. 
Furthermore, the complementary local filtering helps the learning of generalizable feature representation.

\subsection{Comparison with the State-of-the-arts}
\label{section:sota}

\subsubsection{Performances on close-set classification (Task-1)}
In Table \ref{table:OfficeHome}, we show the comparisons with the state-of-the-art approaches for Task-1 on Office-Home dataset. 
All our reported results are averaged over five runs. 
We observe that our proposed \ours~outperforms most existing DG methods and achieves \textbf{67.6\%} classification accuracy on average using ResNet-18, and outperforms the second-best method MixStyle \cite{zhou2021domain} by \textbf{2.2\%} average classification accuracy.  




\subsubsection{Performances on open-set retrieval (Task-2)}
As shown in Table \ref{table:small} and Table \ref{table:big}, our DFF model achieves significant performance improvements on all settings. 
Specifically, (see the Table \ref{table:small}), the mean performance of our method exceeds the second-best by \textbf{8.9\%} in R1 accuracy and \textbf{7.8\%} in mAP. 
As shown in Table \ref{table:big}, our DFF outperforms the second-best by \textbf{3.4\%, 6.7\%, 5.2\%} in R1 accuracy and \textbf{8.0\%, 7.1\%, 7.3\%} in mAP on Market-1501, CUHK03, MSMT17, respectively. When under the setting (\ieno, the testing set of the seen domains are also included for training model), our DFF performs better than the second-best by \textbf{7.3\%, 8.1\%, 5.2\%} in R1 accuracy and \textbf{13.8\%, 8.5\%, 6.5\%} in mAP. 
The results demonstrate that our DFF can significantly improve the generalization ability of learned features even with a simple learned filtering operation in the frequency domain.




\begin{table}[]
\caption{Comparison of Parameters (\#Para), GFLOPs and the Top-1 classification accuracy (Acc.) on ImageNet-1K between the models equipped with \ours~and their corresponding base models. ``\emph{DFF-R18/R50}'' denote the ResNet-18/-50 models equipped with our \ours.}
\label{table:imagenet}
\centering
\scalebox{0.70}{
\begin{tabular}{l|ccc||l|ccc}
\hline
\multicolumn{1}{c|}{Model} & \#Para &  GFLOPs & Acc. & \multicolumn{1}{c|}{Model} & \#Para &  GFLOPs & Acc. \\ \hline
ResNet18                   & 11.7M & 1.8 & 69.8 & ResNet50 & 25.6M & 4.1 & 76.3 \\
\rowcolor{gray!10}DFF-R18  & 12.2M & 2.0 & 72.3 & DFF-R50 & 27.7M & 4.5 & 77.9 \\ \hline
\end{tabular}}
\vspace{-0.5em}
\end{table}

\subsection{Complexity Analysis}
In Table \ref{table:imagenet}, we compare the complexities of our DFF and vanilla ResNet models.
The GFLOPs are calculated with input size of $224 \times 224$. 
FFT and inverse FFT are parameter-free and our filtering design uses average-pooling and max-pooling operations to reduce computational cost. 
Thus, our DFF module only brings limited extra GFLOPs and parameters. 
Our experiment results demonstrate significant performance gain over vanilla ResNet variants. 

\begin{figure}[t]
	\setlength{\abovecaptionskip}{0pt} 
\setlength{\belowcaptionskip}{-0pt}
	\begin{center}
		\includegraphics[width=1.0\linewidth]{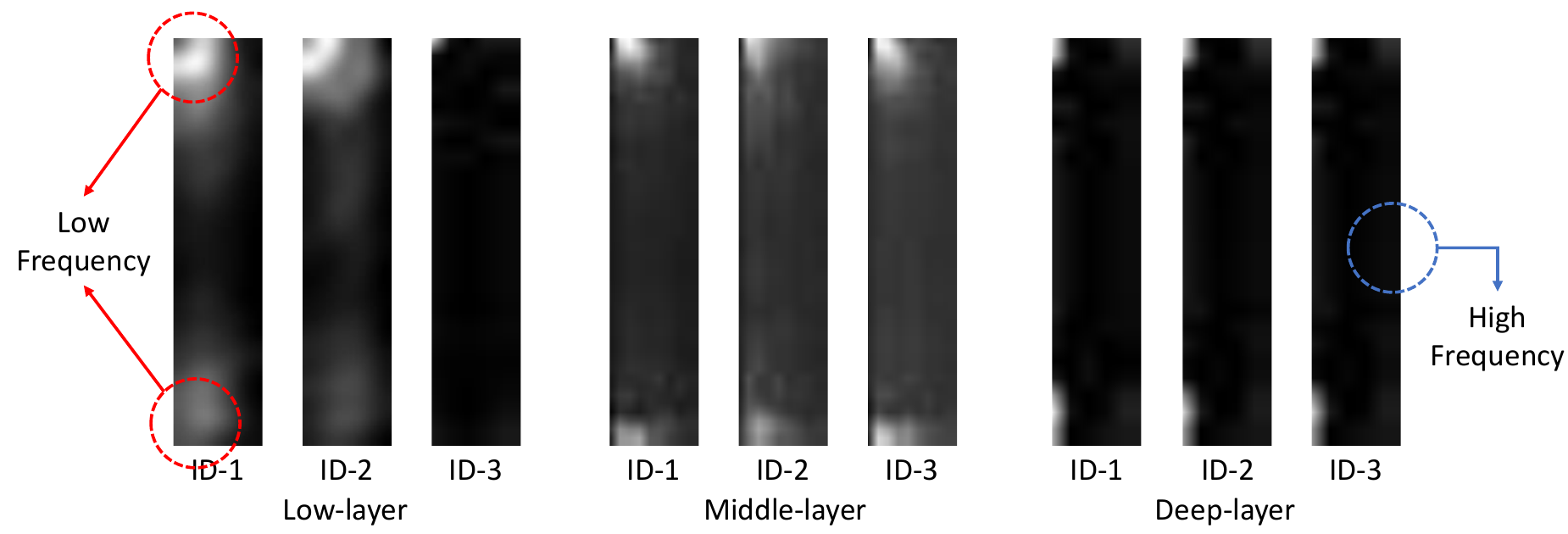}
	\end{center}
	\caption{Visualization of the learned spatial masks of \ours~for the filtering in the frequency domain. The grayscale denotes the mask value, while ``white'' and ``black'' correspond to ``1'' and ``0'', respectively. 
	For each mask, the left top and bottom corners correspond to the low-frequency components while the right middle corresponds to the high-frequency components. The masks at different depths are resized to the same resolution.  }
	\label{fig:mask}
	\vspace{-0.5em}
\end{figure}

\subsection{Visualization of Learned Masks}

We visualize the learned masks at different depths used for \ourfull~in our proposed scheme. The visualization results are in Fig.~\ref{fig:mask}. We draw two observations: 1) The models equipped with \ours~tend to enhance relatively low frequency components while suppressing relatively high ones in the latent frequency domain, which is consistently observed over different instances (IDs). This is in line with the results of theoretical study on the relationship between frequency and robustness \cite{xu2019training,xu2018understanding,wang2020high}. 2) The learned masks are instance-adaptive, demonstrating our proposed scheme can achieve instance-adaptive frequency modulation/filtering in the latent space.

\subsection{Visualization of Learned Feature Maps}

\begin{figure}[h]
	\setlength{\abovecaptionskip}{0pt} 
    \setlength{\belowcaptionskip}{-0pt}
	\begin{center}
		\includegraphics[width=0.8\linewidth]{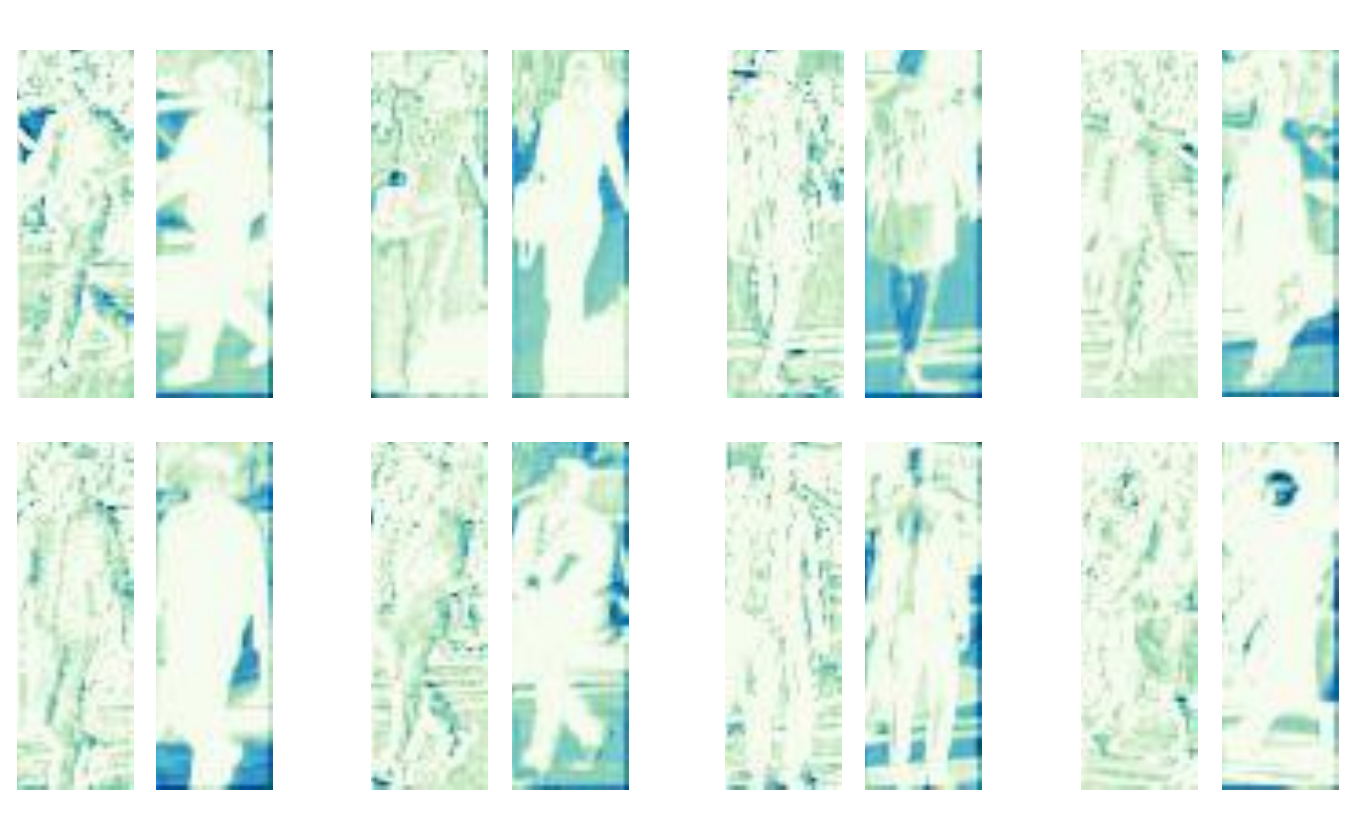}
	\end{center}
 \vspace{-0.3em}
	\caption{We compare the feature maps extracted by the model without (left) and the model with DFF (right). The \textit{lighter} the color is, the \textit{larger} the feature value is.}
	\label{fig:featuremap}
	\vspace{-1.2em}
\end{figure}

We compare the feature maps extracted by the model equipped with our proposed DFF and the one without DFF (see Fig.~\ref{fig:featuremap}). We observe that the features learned by the model equipped with DFF have higher responses for human-body regions than those learned by the baselines model without DFF. This indicates that DFF enables neural networks to focus more precisely on target regions and suppress unrelated feature components (\egno, backgrounds).



\vspace{-0.8em}




Moreover, we also perform t-SNE visualization for the ReID feature vectors learned by the baseline and the model with DFF. The results and analysis are in the supplementary.

\section{Conclusion}

In this paper, we first conceptualize \ourfull~(\ours) and point out that such a simple mechanism can significantly enhance the generalization ability of deep neural networks across domains. This provides a novel alternative for this field. Furthermore, we discuss the implementations of \ours~and showcase the implementation with a simple spatial attention in the frequency domain can bring stunning performance improvements for DG.
Extensive experiments and ablation studies demonstrate the effectiveness of our proposed method. We leave the exploration on more effective instantiations of our conceptualized DFF in the future work, and encourage more combinations and interplay between conventional signal processing and deep learning technologies.




{\small
\bibliographystyle{ieee_fullname}
\bibliography{main}

\begin{thebibliography}{100}\itemsep=-1pt

\bibitem{arnab2021vivit}
Anurag Arnab, Mostafa Dehghani, Georg Heigold, Chen Sun, Mario Lu{\v{c}}i{\'c},
  and Cordelia Schmid.
\newblock Vivit: A video vision transformer.
\newblock In {\em ICCV}, pages 6836--6846, 2021.

\bibitem{bai2021person30k}
Yan Bai, Jile Jiao, Wang Ce, Jun Liu, Yihang Lou, Xuetao Feng, and Ling-Yu
  Duan.
\newblock Person30k: A dual-meta generalization network for person
  re-identification.
\newblock In {\em CVPR}, pages 2123--2132, 2021.

\bibitem{balaji2018metareg}
Yogesh Balaji, Swami Sankaranarayanan, and Rama Chellappa.
\newblock Metareg: Towards domain generalization using meta-regularization.
\newblock {\em NIPS}, 31, 2018.

\bibitem{baxes1994digital}
Gregory~A Baxes.
\newblock {\em Digital image processing: principles and applications}.
\newblock John Wiley \& Sons, Inc., 1994.

\bibitem{bello2019attention}
Irwan Bello, Barret Zoph, Ashish Vaswani, Jonathon Shlens, and Quoc~V Le.
\newblock Attention augmented convolutional networks.
\newblock In {\em ICCV}, pages 3286--3295, 2019.

\bibitem{DBLP:conf/icml/BertasiusWT21}
Gedas Bertasius, Heng Wang, and Lorenzo Torresani.
\newblock Is space-time attention all you need for video understanding?
\newblock In {\em ICML}, pages 813--824, 2021.

\bibitem{carion2020end}
Nicolas Carion, Francisco Massa, Gabriel Synnaeve, Nicolas Usunier, Alexander
  Kirillov, and Sergey Zagoruyko.
\newblock End-to-end object detection with transformers.
\newblock In {\em ECCV}, pages 213--229, 2020.

\bibitem{carlucci2019domain}
Fabio~M Carlucci, Antonio D'Innocente, Silvia Bucci, Barbara Caputo, and
  Tatiana Tommasi.
\newblock Domain generalization by solving jigsaw puzzles.
\newblock In {\em CVPR}, pages 2229--2238, 2019.

\bibitem{chen2021dual}
Peixian Chen, Pingyang Dai, Jianzhuang Liu, Feng Zheng, Qi Tian, and Rongrong
  Ji.
\newblock Dual distribution alignment network for generalizable person
  re-identification.
\newblock In {\em AAAI}, volume~6, 2021.

\bibitem{chen2021explainable}
Xiaodong Chen, Xinchen Liu, Wu Liu, Xiao-Ping Zhang, Yongdong Zhang, and Tao
  Mei.
\newblock Explainable person re-identification with attribute-guided metric
  distillation.
\newblock In {\em ICCV}, pages 11813--11822, 2021.

\bibitem{chi2020fast}
Lu Chi, Borui Jiang, and Yadong Mu.
\newblock Fast fourier convolution.
\newblock In {\em NeurIPS}, pages 4479--4488, 2020.

\bibitem{chi2019fast}
Lu Chi, Guiyu Tian, Yadong Mu, Lingxi Xie, and Qi Tian.
\newblock Fast non-local neural networks with spectral residual learning.
\newblock In {\em Proceedings of the 27th ACM International Conference on
  Multimedia}, pages 2142--2151, 2019.

\bibitem{chitsaz2020acceleration}
Kamran Chitsaz, Mohsen Hajabdollahi, Nader Karimi, Shadrokh Samavi, and Shahram
  Shirani.
\newblock Acceleration of convolutional neural network using fft-based split
  convolutions.
\newblock {\em arXiv preprint arXiv:2003.12621}, 2020.

\bibitem{choi2021meta}
Seokeon Choi, Taekyung Kim, Minki Jeong, Hyoungseob Park, and Changick Kim.
\newblock Meta batch-instance normalization for generalizable person
  re-identification.
\newblock In {\em Proceedings of the IEEE/CVF conference on Computer Vision and
  Pattern Recognition}, pages 3425--3435, 2021.

\bibitem{cooley1965algorithm}
James~W Cooley and John~W Tukey.
\newblock An algorithm for the machine calculation of complex fourier series.
\newblock {\em Mathematics of computation}, 19(90):297--301, 1965.

\bibitem{dai2021generalizable}
Yongxing Dai, Xiaotong Li, Jun Liu, Zekun Tong, and Ling-Yu Duan.
\newblock Generalizable person re-identification with relevance-aware mixture
  of experts.
\newblock In {\em CVPR}, pages 16145--16154, 2021.

\bibitem{dai2021coatnet}
Zihang Dai, Hanxiao Liu, Quoc Le, and Mingxing Tan.
\newblock Coatnet: Marrying convolution and attention for all data sizes.
\newblock In {\em NeurIPS}, volume~34, 2021.

\bibitem{DBLP:conf/iclr/DosovitskiyB0WZ21}
Alexey Dosovitskiy, Lucas Beyer, Alexander Kolesnikov, Dirk Weissenborn,
  Xiaohua Zhai, Thomas Unterthiner, Mostafa Dehghani, Matthias Minderer, Georg
  Heigold, Sylvain Gelly, Jakob Uszkoreit, and Neil Houlsby.
\newblock An image is worth 16x16 words: Transformers for image recognition at
  scale.
\newblock In {\em ICLR}, 2021.

\bibitem{dou2019domain}
Qi Dou, Daniel~C Castro, Konstantinos Kamnitsas, and Ben Glocker.
\newblock Domain generalization via model-agnostic learning of semantic
  features.
\newblock In {\em NeurIPS}, pages 6447--6458, 2019.

\bibitem{d2018domain}
Antonio D’Innocente and Barbara Caputo.
\newblock Domain generalization with domain-specific aggregation modules.
\newblock In {\em GCPR}, pages 187--198. Springer, 2018.

\bibitem{fan2021multiscale}
Haoqi Fan, Bo Xiong, Karttikeya Mangalam, Yanghao Li, Zhicheng Yan, Jitendra
  Malik, and Christoph Feichtenhofer.
\newblock Multiscale vision transformers.
\newblock In {\em ICCV}, pages 6824--6835, 2021.

\bibitem{ganin2015unsupervised}
Yaroslav Ganin and Victor Lempitsky.
\newblock Unsupervised domain adaptation by backpropagation.
\newblock In {\em International conference on machine learning}, pages
  1180--1189. PMLR, 2015.

\bibitem{ghifary2015domain}
Muhammad Ghifary, W~Bastiaan Kleijn, Mengjie Zhang, and David Balduzzi.
\newblock Domain generalization for object recognition with multi-task
  autoencoders.
\newblock In {\em ICCV}, pages 2551--2559, 2015.

\bibitem{gong2019dlow}
Rui Gong, Wen Li, Yuhua Chen, and Luc~Van Gool.
\newblock Dlow: Domain flow for adaptation and generalization.
\newblock In {\em CVPR}, pages 2477--2486, 2019.

\bibitem{gray2008viewpoint}
Douglas Gray and Hai Tao.
\newblock Viewpoint invariant pedestrian recognition with an ensemble of
  localized features.
\newblock In {\em ECCV}, pages 262--275, 2008.

\bibitem{he2016deep}
Kaiming He, Xiangyu Zhang, Shaoqing Ren, and Jian Sun.
\newblock Deep residual learning for image recognition.
\newblock In {\em CVPR}, 2016.

\bibitem{he2020fastreid}
Lingxiao He, Xingyu Liao, Wu Liu, Xinchen Liu, Peng Cheng, and Tao Mei.
\newblock Fastreid: A pytorch toolbox for general instance re-identification.
\newblock {\em arXiv preprint arXiv:2006.02631}, 2020.

\bibitem{he2020guided}
Lingxiao He and Wu Liu.
\newblock Guided saliency feature learning for person re-identification in
  crowded scenes.
\newblock In {\em ECCV}, pages 357--373, 2020.

\bibitem{he2019foreground}
Lingxiao He, Yinggang Wang, Wu Liu, He Zhao, Zhenan Sun, and Jiashi Feng.
\newblock Foreground-aware pyramid reconstruction for alignment-free occluded
  person re-identification.
\newblock In {\em ICCV}, pages 8450--8459, 2019.

\bibitem{hirzer2011person}
Martin Hirzer, Csaba Beleznai, Peter~M Roth, and Horst Bischof.
\newblock Person re-identification by descriptive and discriminative
  classification.
\newblock In {\em SCIA}, pages 91--102, 2011.

\bibitem{honarvar2020domain}
Narges Honarvar~Nazari and Adriana Kovashka.
\newblock Domain generalization using shape representation.
\newblock In {\em ECCV}, pages 666--670, 2020.

\bibitem{hu2018squeeze}
Jie Hu, Li Shen, and Gang Sun.
\newblock Squeeze-and-excitation networks.
\newblock In {\em CVPR}, pages 7132--7141, 2018.

\bibitem{huang2017densely}
Gao Huang, Zhuang Liu, Laurens Van Der~Maaten, and Kilian~Q Weinberger.
\newblock Densely connected convolutional networks.
\newblock In {\em Proceedings of the IEEE conference on computer vision and
  pattern recognition}, pages 4700--4708, 2017.

\bibitem{huang2021fsdr}
Jiaxing Huang, Dayan Guan, Aoran Xiao, and Shijian Lu.
\newblock Fsdr: Frequency space domain randomization for domain generalization.
\newblock In {\em CVPR}, pages 6891--6902, 2021.

\bibitem{huang2020self}
Zeyi Huang, Haohan Wang, Eric~P Xing, and Dong Huang.
\newblock Self-challenging improves cross-domain generalization.
\newblock In {\em ECCV}, pages 124--140. Springer, 2020.

\bibitem{jaderberg2015spatial}
Max Jaderberg, Karen Simonyan, Andrew Zisserman, et~al.
\newblock Spatial transformer networks.
\newblock {\em NeurIPS}, 28, 2015.

\bibitem{jin2020style}
Xin Jin, Cuiling Lan, Wenjun Zeng, Zhibo Chen, and Li Zhang.
\newblock Style normalization and restitution for generalizable person
  re-identification.
\newblock In {\em CVPR}, pages 3143--3152, 2020.

\bibitem{katznelson2004introduction}
Yitzhak Katznelson.
\newblock {\em An introduction to harmonic analysis}.
\newblock Cambridge University Press, 2004.

\bibitem{kingma2014adam}
Diederik~P Kingma and Jimmy Ba.
\newblock Adam: A method for stochastic optimization.
\newblock {\em arXiv preprint arXiv:1412.6980}, 2014.

\bibitem{li2018learning}
Da Li, Yongxin Yang, Yi-Zhe Song, and Timothy Hospedales.
\newblock Learning to generalize: Meta-learning for domain generalization.
\newblock In {\em AAAI}, volume~32, 2018.

\bibitem{li2017deeper}
Da Li, Yongxin Yang, Yi-Zhe Song, and Timothy~M Hospedales.
\newblock Deeper, broader and artier domain generalization.
\newblock In {\em ICCV}, pages 5542--5550, 2017.

\bibitem{li2019episodic}
Da Li, Jianshu Zhang, Yongxin Yang, Cong Liu, Yi-Zhe Song, and Timothy~M
  Hospedales.
\newblock Episodic training for domain generalization.
\newblock In {\em ICCV}, pages 1446--1455, 2019.

\bibitem{li2018domain}
Haoliang Li, Sinno~Jialin Pan, Shiqi Wang, and Alex~C Kot.
\newblock Domain generalization with adversarial feature learning.
\newblock In {\em CVPR}, pages 5400--5409, 2018.

\bibitem{li2021combined}
Hanjun Li, Gaojie Wu, and Wei-Shi Zheng.
\newblock Combined depth space based architecture search for person
  re-identification.
\newblock In {\em CVPR}, pages 6729--6738, 2021.

\bibitem{li2020falcon}
Shaohua Li, Kaiping Xue, Bin Zhu, Chenkai Ding, Xindi Gao, David Wei, and Tao
  Wan.
\newblock Falcon: A fourier transform based approach for fast and secure
  convolutional neural network predictions.
\newblock In {\em CVPR}, pages 8705--8714, 2020.

\bibitem{li2013locally}
Wei Li and Xiaogang Wang.
\newblock Locally aligned feature transforms across views.
\newblock In {\em CVPR}, pages 3594--3601, 2013.

\bibitem{li2014deepreid}
Wei Li, Rui Zhao, Tong Xiao, and Xiaogang Wang.
\newblock Deepreid: Deep filter pairing neural network for person
  re-identification.
\newblock In {\em CVPR}, pages 152--159, 2014.

\bibitem{DBLP:journals/corr/abs-2111-13420}
Xin Li, Zhizheng Zhang, Guoqiang Wei, Cuiling Lan, Wenjun Zeng, Xin Jin, and
  Zhibo Chen.
\newblock Confounder identification-free causal visual feature learning.
\newblock {\em CoRR}, abs/2111.13420, 2021.

\bibitem{li2018deep}
Ya Li, Xinmei Tian, Mingming Gong, Yajing Liu, Tongliang Liu, Kun Zhang, and
  Dacheng Tao.
\newblock Deep domain generalization via conditional invariant adversarial
  networks.
\newblock In {\em ECCV}, pages 624--639, 2018.

\bibitem{li2019feature}
Yiying Li, Yongxin Yang, Wei Zhou, and Timothy Hospedales.
\newblock Feature-critic networks for heterogeneous domain generalization.
\newblock In {\em ICML}, pages 3915--3924. PMLR, 2019.

\bibitem{liao2020interpretable}
Shengcai Liao and Ling Shao.
\newblock Interpretable and generalizable person re-identification with
  query-adaptive convolution and temporal lifting.
\newblock In {\em ECCV}, pages 456--474. Springer, 2020.

\bibitem{lin2021domain}
Ci-Siang Lin, Yuan-Chia Cheng, and Yu-Chiang~Frank Wang.
\newblock Domain generalized person re-identification via cross-domain episodic
  learning.
\newblock In {\em ICPR}, pages 6758--6763. IEEE, 2021.

\bibitem{liu2022debiased}
Jiawei Liu, Zhipeng Huang, Liang Li, Kecheng Zheng, and Zheng-Jun Zha.
\newblock Debiased batch normalization via gaussian process for generalizable
  person re-identification.
\newblock In {\em AAAI}, 2022.

\bibitem{liu2021feddg}
Quande Liu, Cheng Chen, Jing Qin, Qi Dou, and Pheng-Ann Heng.
\newblock Feddg: Federated domain generalization on medical image segmentation
  via episodic learning in continuous frequency space.
\newblock In {\em CVPR}, pages 1013--1023, 2021.

\bibitem{loy2009multi}
Chen~Change Loy, Tao Xiang, and Shaogang Gong.
\newblock Multi-camera activity correlation analysis.
\newblock In {\em Proceedings of the IEEE Conference on Computer Vision and
  Pattern Recognition}, pages 1988--1995. IEEE, 2009.

\bibitem{luo2019bag}
Hao Luo, Youzhi Gu, Xingyu Liao, Shenqi Lai, and Wei Jiang.
\newblock Bag of tricks and a strong baseline for deep person
  re-identification.
\newblock In {\em CVPR Workshops}, pages 0--0, 2019.

\bibitem{mancini2018best}
Massimiliano Mancini, Samuel~Rota Bulo, Barbara Caputo, and Elisa Ricci.
\newblock Best sources forward: domain generalization through source-specific
  nets.
\newblock In {\em ICIP}, pages 1353--1357, 2018.

\bibitem{mao2021deep}
Xintian Mao, Yiming Liu, Wei Shen, Qingli Li, and Yan Wang.
\newblock Deep residual fourier transformation for single image deblurring.
\newblock {\em arXiv preprint arXiv:2111.11745}, 2021.

\bibitem{mathieu2013fast}
Michael Mathieu, Mikael Henaff, and Yann LeCun.
\newblock Fast training of convolutional networks through ffts.
\newblock {\em arXiv preprint arXiv:1312.5851}, 2013.

\bibitem{matsuura2020domain}
Toshihiko Matsuura and Tatsuya Harada.
\newblock Domain generalization using a mixture of multiple latent domains.
\newblock In {\em AAAI}, volume~34, pages 11749--11756, 2020.

\bibitem{misra2021rotate}
Diganta Misra, Trikay Nalamada, Ajay~Uppili Arasanipalai, and Qibin Hou.
\newblock Rotate to attend: Convolutional triplet attention module.
\newblock In {\em WACV}, pages 3139--3148, 2021.

\bibitem{motiian2017unified}
Saeid Motiian, Marco Piccirilli, Donald~A Adjeroh, and Gianfranco Doretto.
\newblock Unified deep supervised domain adaptation and generalization.
\newblock In {\em ICCV}, pages 5715--5725, 2017.

\bibitem{nair2020fast}
Varsha Nair, Moitrayee Chatterjee, Neda Tavakoli, Akbar~Siami Namin, and Craig
  Snoeyink.
\newblock Fast fourier transformation for optimizing convolutional neural
  networks in object recognition.
\newblock {\em arXiv preprint arXiv:2010.04257}, 2020.

\bibitem{pan2018two}
Xingang Pan, Ping Luo, Jianping Shi, and Xiaoou Tang.
\newblock Two at once: Enhancing learning and generalization capacities via
  ibn-net.
\newblock In {\em ECCV}, 2018.

\bibitem{patrick2021keeping}
Mandela Patrick, Dylan Campbell, Yuki Asano, Ishan Misra, Florian Metze,
  Christoph Feichtenhofer, Andrea Vedaldi, and Jo{\~a}o~F Henriques.
\newblock Keeping your eye on the ball: Trajectory attention in video
  transformers.
\newblock In {\em NeurIPS}, 2021.

\bibitem{pitas2000digital}
Ioannis Pitas.
\newblock {\em Digital image processing algorithms and applications}.
\newblock John Wiley \& Sons, 2000.

\bibitem{prabhu2020butterfly}
Anish Prabhu, Ali Farhadi, Mohammad Rastegari, et~al.
\newblock Butterfly transform: An efficient fft based neural architecture
  design.
\newblock In {\em CVPR}, pages 12024--12033, 2020.

\bibitem{pratt2017fcnn}
Harry Pratt, Bryan Williams, Frans Coenen, and Yalin Zheng.
\newblock Fcnn: Fourier convolutional neural networks.
\newblock In {\em ECML PKDD}, pages 786--798, 2017.

\bibitem{qin2021fcanet}
Zequn Qin, Pengyi Zhang, Fei Wu, and Xi Li.
\newblock Fcanet: Frequency channel attention networks.
\newblock In {\em ICCV}, pages 783--792, 2021.

\bibitem{rao2021global}
Yongming Rao, Wenliang Zhao, Zheng Zhu, Jiwen Lu, and Jie Zhou.
\newblock Global filter networks for image classification.
\newblock In {\em NeurIPS}, volume~34, 2021.

\bibitem{russakovsky2015imagenet}
Olga Russakovsky, Jia Deng, Hao Su, Jonathan Krause, Sanjeev Satheesh, Sean Ma,
  Zhiheng Huang, Andrej Karpathy, Aditya Khosla, Michael Bernstein, et~al.
\newblock Imagenet large scale visual recognition challenge.
\newblock {\em International journal of computer vision}, 115(3):211--252,
  2015.

\bibitem{seo2020learning}
Seonguk Seo, Yumin Suh, Dongwan Kim, Geeho Kim, Jongwoo Han, and Bohyung Han.
\newblock Learning to optimize domain specific normalization for domain
  generalization.
\newblock In {\em ECCV}, pages 68--83, 2020.

\bibitem{shankar2018generalizing}
Shiv Shankar, Vihari Piratla, Soumen Chakrabarti, Siddhartha Chaudhuri, Preethi
  Jyothi, and Sunita Sarawagi.
\newblock Generalizing across domains via cross-gradient training.
\newblock {\em ICLR}, 2018.

\bibitem{song2019generalizable}
Jifei Song, Yongxin Yang, Yi-Zhe Song, Tao Xiang, and Timothy~M Hospedales.
\newblock Generalizable person re-identification by domain-invariant mapping
  network.
\newblock In {\em CVPR}, pages 719--728, 2019.

\bibitem{sun2020circle}
Yifan Sun, Changmao Cheng, Yuhan Zhang, Chi Zhang, Liang Zheng, Zhongdao Wang,
  and Yichen Wei.
\newblock Circle loss: A unified perspective of pair similarity optimization.
\newblock In {\em CVPR}, pages 6398--6407, 2020.

\bibitem{sun2018beyond}
Yifan Sun, Liang Zheng, Yi Yang, Qi Tian, and Shengjin Wang.
\newblock Beyond part models: Person retrieval with refined part pooling (and a
  strong convolutional baseline).
\newblock In {\em ECCV}, pages 480--496, 2018.

\bibitem{suvorov2022resolution}
Roman Suvorov, Elizaveta Logacheva, Anton Mashikhin, Anastasia Remizova,
  Arsenii Ashukha, Aleksei Silvestrov, Naejin Kong, Harshith Goka, Kiwoong
  Park, and Victor Lempitsky.
\newblock Resolution-robust large mask inpainting with fourier convolutions.
\newblock In {\em WACV}, pages 2149--2159, 2022.

\bibitem{tremblay2018training}
Jonathan Tremblay, Aayush Prakash, David Acuna, Mark Brophy, Varun Jampani, Cem
  Anil, Thang To, Eric Cameracci, Shaad Boochoon, and Stan Birchfield.
\newblock Training deep networks with synthetic data: Bridging the reality gap
  by domain randomization.
\newblock In {\em CVPR workshops}, pages 969--977, 2018.

\bibitem{van2008visualizing}
Laurens Van~der Maaten and Geoffrey Hinton.
\newblock Visualizing data using t-sne.
\newblock {\em Journal of machine learning research}, 9(11), 2008.

\bibitem{venkateswara2017deep}
Hemanth Venkateswara, Jose Eusebio, Shayok Chakraborty, and Sethuraman
  Panchanathan.
\newblock Deep hashing network for unsupervised domain adaptation.
\newblock In {\em CVPR}, pages 5018--5027, 2017.

\bibitem{volpi2018generalizing}
Riccardo Volpi, Hongseok Namkoong, Ozan Sener, John~C Duchi, Vittorio Murino,
  and Silvio Savarese.
\newblock Generalizing to unseen domains via adversarial data augmentation.
\newblock {\em NeurIPS}, 31, 2018.

\bibitem{wang2018learning}
Guanshuo Wang, Yufeng Yuan, Xiong Chen, Jiwei Li, and Xi Zhou.
\newblock Learning discriminative features with multiple granularities for
  person re-identification.
\newblock In {\em ACMMM}, pages 274--282, 2018.

\bibitem{wang2020high}
Haohan Wang, Xindi Wu, Zeyi Huang, and Eric~P Xing.
\newblock High-frequency component helps explain the generalization of
  convolutional neural networks.
\newblock In {\em CVPR}, 2020.

\bibitem{DBLP:conf/cvpr/WangWZLZH20}
Qilong Wang, Banggu Wu, Pengfei Zhu, Peihua Li, Wangmeng Zuo, and Qinghua Hu.
\newblock Eca-net: Efficient channel attention for deep convolutional neural
  networks.
\newblock In {\em CVPR}, pages 11531--11539, 2020.

\bibitem{wang2018non}
Xiaolong Wang, Ross Girshick, Abhinav Gupta, and Kaiming He.
\newblock Non-local neural networks.
\newblock In {\em CVPR}, pages 7794--7803, 2018.

\bibitem{wei2021toalign}
Guoqiang Wei, Cuiling Lan, Wenjun Zeng, Zhizheng Zhang, and Zhibo Chen.
\newblock Toalign: task-oriented alignment for unsupervised domain adaptation.
\newblock {\em NeurIPS}, 34:13834--13846, 2021.

\bibitem{wei2018person}
Longhui Wei, Shiliang Zhang, Wen Gao, and Qi Tian.
\newblock Person transfer gan to bridge domain gap for person
  re-identification.
\newblock In {\em CVPR}, pages 79--88, 2018.

\bibitem{woo2018cbam}
Sanghyun Woo, Jongchan Park, Joon-Young Lee, and In~So Kweon.
\newblock Cbam: Convolutional block attention module.
\newblock In {\em ECCV}, pages 3--19, 2018.

\bibitem{xiao2017joint}
Tong Xiao, Shuang Li, Bochao Wang, Liang Lin, and Xiaogang Wang.
\newblock Joint detection and identification feature learning for person
  search.
\newblock In {\em CVPR}, pages 3415--3424, 2017.

\bibitem{xie2017aggregated}
Saining Xie, Ross Girshick, Piotr Doll{\'a}r, Zhuowen Tu, and Kaiming He.
\newblock Aggregated residual transformations for deep neural networks.
\newblock In {\em Proceedings of the IEEE conference on computer vision and
  pattern recognition}, pages 1492--1500, 2017.

\bibitem{xu2021fourier}
Qinwei Xu, Ruipeng Zhang, Ya Zhang, Yanfeng Wang, and Qi Tian.
\newblock A fourier-based framework for domain generalization.
\newblock In {\em CVPR}, pages 14383--14392, 2021.

\bibitem{xu2018understanding}
Zhiqin~John Xu.
\newblock Understanding training and generalization in deep learning by fourier
  analysis.
\newblock {\em arXiv preprint arXiv:1808.04295}, 2018.

\bibitem{xu2019training}
Zhi-Qin~John Xu, Yaoyu Zhang, and Yanyang Xiao.
\newblock Training behavior of deep neural network in frequency domain.
\newblock In {\em ICONIP}, 2019.

\bibitem{yan2021beyond}
Cheng Yan, Guansong Pang, Xiao Bai, Changhong Liu, Ning Xin, Lin Gu, and Jun
  Zhou.
\newblock Beyond triplet loss: person re-identification with fine-grained
  difference-aware pairwise loss.
\newblock {\em IEEE Trans Multimedia}, 2021.

\bibitem{yang2020fda}
Yanchao Yang and Stefano Soatto.
\newblock Fda: Fourier domain adaptation for semantic segmentation.
\newblock In {\em CVPR}, pages 4085--4095, 2020.

\bibitem{yi2021contrastive}
Qiaosi Yi, Jinhao Liu, Le Hu, Faming Fang, and Guixu Zhang.
\newblock Contrastive learning for local and global learning mri
  reconstruction.
\newblock {\em arXiv preprint arXiv:2111.15200}, 2021.

\bibitem{yin2019fourier}
Dong Yin, Raphael Gontijo~Lopes, Jon Shlens, Ekin~Dogus Cubuk, and Justin
  Gilmer.
\newblock A fourier perspective on model robustness in computer vision.
\newblock {\em NeurIPS}, 2019.

\bibitem{yue2019domain}
Xiangyu Yue, Yang Zhang, Sicheng Zhao, Alberto Sangiovanni-Vincentelli, Kurt
  Keutzer, and Boqing Gong.
\newblock Domain randomization and pyramid consistency: Simulation-to-real
  generalization without accessing target domain data.
\newblock In {\em ICCV}, pages 2100--2110, 2019.

\bibitem{zhang2020resnest}
Hang Zhang, Chongruo Wu, Zhongyue Zhang, Yi Zhu, Haibin Lin, Zhi Zhang, Yue
  Sun, Tong He, Jonas Mueller, R Manmatha, et~al.
\newblock Resnest: Split-attention networks.
\newblock {\em arXiv preprint arXiv:2004.08955}, 2020.

\bibitem{zhang2021learning}
Yi-Fan Zhang, Hanlin Zhang, Zhang Zhang, Da Li, Zhen Jia, Liang Wang, and
  Tieniu Tan.
\newblock Learning domain invariant representations for generalizable person
  re-identification.
\newblock {\em arXiv preprint arXiv:2103.15890}, 2021.

\bibitem{zhang2019densely}
Zhizheng Zhang, Cuiling Lan, Wenjun Zeng, and Zhibo Chen.
\newblock Densely semantically aligned person re-identification.
\newblock In {\em CVPR}, pages 667--676, 2019.

\bibitem{zhang2020multi}
Zhizheng Zhang, Cuiling Lan, Wenjun Zeng, and Zhibo Chen.
\newblock Multi-granularity reference-aided attentive feature aggregation for
  video-based person re-identification.
\newblock In {\em CVPR}, pages 10407--10416, 2020.

\bibitem{zhangbeyond}
Zhizheng Zhang, Cuiling Lan, Wenjun Zeng, Zhibo Chen, and Shih-Fu Chang.
\newblock Beyond triplet loss: Meta prototypical n-tuple loss for person
  re-identification.
\newblock {\em IEEE Trans Multimedia}, 2021.

\bibitem{zhang2020relation}
Zhizheng Zhang, Cuiling Lan, Wenjun Zeng, Xin Jin, and Zhibo Chen.
\newblock Relation-aware global attention for person re-identification.
\newblock In {\em CVPR}, pages 3186--3195, 2020.

\bibitem{zhao2020deep}
Cairong Zhao, Xinbi Lv, Zhang Zhang, Wangmeng Zuo, Jun Wu, and Duoqian Miao.
\newblock Deep fusion feature representation learning with hard mining
  center-triplet loss for person re-identification.
\newblock {\em IEEE Trans Multimedia}, pages 3180--3195, 2020.

\bibitem{zhao2020learning}
Yuyang Zhao, Zhun Zhong, Fengxiang Yang, Zhiming Luo, Yaojin Lin, Shaozi Li,
  and Nicu Sebe.
\newblock Learning to generalize unseen domains via memory-based multi-source
  meta-learning for person re-identification.
\newblock In {\em CVPR}, 2021.

\bibitem{zhao2021learning}
Yuyang Zhao, Zhun Zhong, Fengxiang Yang, Zhiming Luo, Yaojin Lin, Shaozi Li,
  and Nicu Sebe.
\newblock Learning to generalize unseen domains via memory-based multi-source
  meta-learning for person re-identification.
\newblock In {\em CVPR}, pages 6277--6286, 2021.

\bibitem{zheng2015scalable}
Liang Zheng, Liyue Shen, Lu Tian, Shengjin Wang, Jingdong Wang, and Qi Tian.
\newblock Scalable person re-identification: A benchmark.
\newblock In {\em ICCV}, pages 1116--1124, 2015.

\bibitem{zheng2009associating}
Wei-Shi Zheng, Shaogang Gong, and Tao Xiang.
\newblock Associating groups of people.
\newblock In {\em BMVC}, volume~2, pages 1--11, 2009.

\bibitem{zheng2019joint}
Zhedong Zheng, Xiaodong Yang, Zhiding Yu, Liang Zheng, Yi Yang, and Jan Kautz.
\newblock Joint discriminative and generative learning for person
  re-identification.
\newblock In {\em CVPR}, pages 2138--2147, 2019.

\bibitem{zheng2017unlabeled}
Zhedong Zheng, Liang Zheng, and Yi Yang.
\newblock Unlabeled samples generated by gan improve the person
  re-identification baseline in vitro.
\newblock In {\em Proceedings of the IEEE international conference on computer
  vision}, pages 3754--3762, 2017.

\bibitem{zhou2019omni}
Kaiyang Zhou, Yongxin Yang, Andrea Cavallaro, and Tao Xiang.
\newblock Omni-scale feature learning for person re-identification.
\newblock In {\em ICCV}, pages 3702--3712, 2019.

\bibitem{zhou2020deep}
Kaiyang Zhou, Yongxin Yang, Timothy Hospedales, and Tao Xiang.
\newblock Deep domain-adversarial image generation for domain generalisation.
\newblock In {\em AAAI}, pages 13025--13032, 2020.

\bibitem{zhou2021domain}
Kaiyang Zhou, Yongxin Yang, Yu Qiao, and Tao Xiang.
\newblock Domain adaptive ensemble learning.
\newblock {\em TIP}, 2021.

\bibitem{zhu2022hmfca}
Ying Zhu, Runwei Ding, Weibo Huang, Peng Wei, Ge Yang, and Yong Wang.
\newblock Hmfca-net: Hierarchical multi-frequency based channel attention net
  for mobile phone surface defect detection.
\newblock {\em Pattern Recognit Lett}, 153:118--125, 2022.

\bibitem{zhuang2020rethinking}
Zijie Zhuang, Longhui Wei, Lingxi Xie, Tianyu Zhang, Hengheng Zhang, Haozhe Wu,
  Haizhou Ai, and Qi Tian.
\newblock Rethinking the distribution gap of person re-identification with
  camera-based batch normalization.
\newblock In {\em ECCV}, pages 140--157. Springer, 2020.

\end{thebibliography}
}

\clearpage
\section*{\Large{\textbf{Supplementary Material}}}

\section{More Datasets and Implementation Details}

We evaluate the effectiveness of our proposed Deep Frequency Filtering (DFF) for Domain Generalization (DG) on \textbf{Task-1}: the close-set classification task and \textbf{Task-2}: the open-set retrieval task, \ieno, person re-identification (ReID). More details about the datasets and our experiment configurations are introduced in this section.

\subsection{Datasets for Task-1}

We use the most commonly used Office-Home \cite{li2017deeper} and PACS dataset~\cite{li2017deeper}. Specifically, Office-Home consists of 4 domains (Art (Ar), Clip Art (Cl), Product (Pr), Real-World (Rw)), each consisting of 65 categories, with an average of 70 images per category, for a total of 15,500 images. PACS consists of 9991 samples in total from 4 domains (\ieno, Photo (P), Art Painting (A), Cartoon (C) and Sketch (S)). All these 4 domains share 7 object categories. 
They are commonly used domain generalization (DG) benchmark on the task of classification. 
We validate the effectiveness of our proposed method for generalization in close-set classification task on Office-Home and PACS.   
Following the typical setting, we conduct experiments on this dataset under the leave-one-out protocol (see Table \ref{table:Protocol} Protocol-1 and Protocol-2), where three domains are used for training and the remaining one is considered as the unknown target domain.

\subsection{Datasets for Task-2}
Person re-identification (ReID) is a representative open-set retrieval task, where different domains and datasets do not share their label space. 
We employ existing ReID protocols to evaluate the generalization ability of our method.
$i)$ For Protocol-3 and Protocol-4, we also follow the leave-one-out protocols as in~\cite{zhao2020learning,liu2022debiased}. 
Among the four datasets (CUHK-SYSU (CS) \cite{xiao2017joint}, MSMT17 (MS) \cite{wei2018person}, CUHK03 (C3) \cite{li2014deepreid} and Market-1501 (MA) \cite{zheng2015scalable}), three are selected as the seen domain for training and the remaining one is selected the unseen domain data for testing. 
Differently, Protocol-3 only adopts the training set of seen
domains for model training while in Protocol-3, the testing set of the seen domains are also included for training model. 
$ii)$ For Protocol-5 in Table \ref{table:Protocol}, several large-scale ReID datasets \egno, CUHK02 (C2)\cite{li2013locally}, CUHK03 (C3) \cite{li2014deepreid}, Market-1501 (MA) \cite{zheng2015scalable} and CUHK-SYSU (CS) \cite{xiao2017joint}, are viewed as multiple source domains. 
Each small-scale ReID dataset including VIPeR \cite{gray2008viewpoint}, PRID \cite{hirzer2011person}, GRID \cite{loy2009multi} and iLIDS \cite{zheng2009associating} is used as an unseen target domain, respectively. 
To comply with the General Ethical Conduct, we exclude DukeMTMC from the source domains. 
The final performance is obtained by averaging $10$ repeated experiments with random splits of training and testing sets.

\begin{table}[t]

	\begin{center}
	\caption{The evaluation protocols. ``Com-'' refers to combining the train and test sets of source domains for training. ``Pr'', ``Ar'', ``Cl'', ``Rw'' are short for the Product, Art, Clip Art and Real-World domains in Office-Home dataset \cite{li2017deeper}, respectively. ``P'', ``A'', ``C'', ``S'' are short for the Photo, Painting, Cartoon, Sketch domains in PACS dataset \cite{li2017deeper}, respectively. ``MA'', ``CS'', ``C3'', ``MS'' denote Market-1501 \cite{zheng2015scalable}, CUHK-SYSU \cite{xiao2017joint}, CUHK03 \cite{li2014deepreid}, MSMT17 \cite{wei2018person}, respectively. Note that for person ReID, the commonly used DukeMTMC \cite{zheng2017unlabeled} has been withdrawn by its publisher, is thus no longer used. }
		\label{table:Protocol}
\scalebox{0.73}{
		\begin{tabular}{l|c|c|c}
		   \hline
		    \multicolumn{1}{c|}{Task} & \multicolumn{1}{c|}{Setting} & \multicolumn{1}{c|}{Training Data} & \multicolumn{1}{c}{Testing Data} \\
		    \hline
            \multirow{8}[0]{*}{Close-set classification} &
            \multirow{4}[0]{*}{Protocol-1} & Cl,Pr,Rw & Ar \\
                  &       & Ar,Pr,Rw & Cl \\
                  &       & Ar,Cl,Rw & Pr  \\
                  &       & Ar,Cl,Pr & Rw \\
                  \cline{2-4} 
            & \multirow{4}[0]{*}{Protocol-2} & C,P,S & A \\
                  &       & A,P,S & C \\
                  &       & A,C,P & S \\
                  &       & A,C,S & P \\
        
            \hline
            \multirow{10}[0]{*}{Open-set retrieval} &\multirow{3}[0]{*}{Protocol-3} 
            & CS+C3+MS & MA \\
                  &       & MA+CS+MS & C3 \\
                  &       & MA+CS+C3 & MS \\ 
            \cline{2-4} 
                  & \multirow{3}[0]{*}{Protocol-4} & Com-(CS+C3+MT) & MA  \\
                  &       & Com-(MA+CS+MS) & C3 \\
                  &       & Com-(MA+CS+C3) & MS \\
                  \cline{2-4} 
                &  \multirow{4}[0]{*}{Protocol-5}  & \multirow{4}[0]{*}{Com-(MA+C2+C3+CS)} & PRID \\
                & & & GRID \\
                & & & VIPeR \\
                & & & iLIDs \\
        \hline
		\end{tabular}%
		}
		\end{center}
	\vspace{-0.50cm}
\end{table}%

\subsection{Networks} 
Following the common practices of domain generalizable classification (Task-1) \cite{li2018domain,shankar2018generalizing,carlucci2019domain,zhou2021domain} and person ReID (Task-2) \cite{choi2021meta,jin2020style,liao2020interpretable,chen2021dual}, we build the networks equipped with our proposed \ourfull~(\ours) for these two tasks on the basis of ResNet-18 and ResNet-50, respectively.
As introduced in the Sec. 3.4 of our manuscript, we evaluate the effectiveness of our proposed \ours~based on the two-branch architecture of Fast Fourier Convolution (FFC) in \cite{chi2020fast}. In particular, we adopt our \ours~operation to the spectral transformer branch of this architecture. Unless otherwise stated, the ratio $r$ in splitting $\mathbf{X} \in \mathbb{R}^{C \times H \times W}$ into $\mathbf{X}^{g} \in \mathbb{R}^{rC \times H \times W}$ and $\mathbf{X}^{l} \in \mathbb{R}^{(1-r)C \times H \times W}$ is set to 0.5. We conduct an ablation study on this ratio in this Supplementary as follows.

\subsection{Training}

Following common practices \cite{carlucci2019domain,matsuura2020domain,li2018domain,zhang2019densely,luo2019bag,he2020fastreid}, we adopt ResNet-18 and ResNet-50 \cite{he2016deep} as our backbone for Task-1 and Task-2, respectively.  
Each convolution layer of the backbone is replaced with our DDF module. 
Unless specially stated, we first pretrain the models on ImageNet \cite{russakovsky2015imagenet} then fine-tune them on Task-1 or Task-2, referring to the common practices \cite{shankar2018generalizing,carlucci2019domain,zhou2021domain,choi2021meta,chen2021dual}. We introduce our training configurations with more details in the following.

\paragraph{Pre-training on ImageNet.}
Following the common practices \cite{he2016deep,xie2017aggregated,huang2017densely,zhang2020resnest} in this field, we adopt the commonly used data augmentation strategies including color jittering, random flipping and center cropping. The input image size is $224 \times 224$. We use the SGD optimizer with the base initial learning rate of 0.4, the momentum of 0.9 and the weight decay of 0.0001, and perform learning rate decay by a factor of 0.1 after 30, 60 and 80 epochs. A linear warm-up strategy is adopted in the first 5 epochs, where the learning rate is increased from 0.0 to 0.4 linearly. All models are trained for 90 epochs with the batchsize of 1024. 

\paragraph{Fine-tuning on Task-1.}
The initial learning rate in this stage is set to 0.001. We train all models on this task using the SGD optimizer with the momentum of 0.9 and the weight decay of 0.0001. The batch size is set to 32. Following prior works \cite{motiian2017unified,shankar2018generalizing,carlucci2019domain,li2018domain}, we adopt the data augmentation strategies including random cropping, horizontal flipping, and random grayscale. The input images are resized to $224 \times 224$. On the PACS dataset \cite{li2017deeper}, we train the models for 3,500 iterations; and on the Office-Home \cite{venkateswara2017deep} dataset, we train the models for 3,000 iterations. The experiment results on Office-Home have been presented in the main paper while the results on PACS are placed in this Supplementary due to the length limitation.

\paragraph{Fine-tuning on Task-2.}
Following the common practices for domain generalizable person ReID \cite{jin2020style,dai2021generalizable,choi2021meta,zhao2021learning}, we adopt the widely used data augmentation strategies, including cropping, random flipping, and color jittering. We use Adam~\cite{kingma2014adam} optimizer with the momentum of 0.9 and weight decay of 0.0005. The learning rate is initialized by $3.5\!\times\!10^{-4}$ and decayed using a cosine annealing schedule. The batch size is set to 128, including 8 identities and 16 images per identity. For the Protocol-3, Protocol-4 and Protocol-5, the models are trained for 60 epochs on their corresponding source datasets.
Similar to previous work~\cite{luo2019bag}, the last spatial down-sampling in the ``conv5\_x'' block is removed. The input images are resized to $384\times128$. Following \cite{he2020fastreid}, we use task-related loss including cross-entropy loss, arcface loss, circle loss and triplet loss. And we adopt a gradient reversal layer \cite{ganin2015unsupervised} encouraging the learning of domain-invariant features.

\section{More Experiment Results}

\begin{table}[!t]
\begin{center}

\caption{Performance (classification accuracy \%) comparison with the state-of-the-art methods under Protocol-2 (\ieno, on PACS dataset) on close-set classification task. We use ResNet-18 as backbone. Best in bold.}
\label{table:PACS}
\scalebox{0.76}{
\begin{tabular}{c|cccc|c}
\hline
\multirow{2}{*}{Method} &  \multicolumn{4}{c|}{Source$\rightarrow$Target}                                                                                                                          & \multirow{2}{*}{Avg} \\ \cline{2-5}                     & \multicolumn{1}{c|}{C,P,S$\rightarrow$A} & \multicolumn{1}{c|}{A,P,S$\rightarrow$C} & \multicolumn{1}{c|}{A,C,P$\rightarrow$S} & A,C,S$\rightarrow$P &                      \\ \hline
\multicolumn{1}{l|}{Baseline }                                    & \multicolumn{1}{c|}{77.6}                  & \multicolumn{1}{c|}{73.9}                  & \multicolumn{1}{c|}{70.3}                  & 94.4                  & 79.1                 \\
\multicolumn{1}{l|}{MMD-AAE \cite{li2018domain}}                              & \multicolumn{1}{c|}{75.2}                  & \multicolumn{1}{c|}{72.7}                  & \multicolumn{1}{c|}{64.2}                  & 96.0                  & 77.0                 \\
\multicolumn{1}{l|}{CrossGrad \cite{shankar2018generalizing}}                            & \multicolumn{1}{c|}{78.7}                  & \multicolumn{1}{c|}{73.3}                  & \multicolumn{1}{c|}{65.1}                  & 94.0                  & 77.8                 \\
\multicolumn{1}{l|}{MetaReg \cite{balaji2018metareg} }                               & \multicolumn{1}{c|}{79.5}                  & \multicolumn{1}{c|}{75.4}                  & \multicolumn{1}{c|}{72.2}                  & 94.3                  & 80.4                 \\
\multicolumn{1}{l|}{JiGen \cite{carlucci2019domain}  }                              & \multicolumn{1}{c|}{79.4}                  & \multicolumn{1}{c|}{77.3}                  & \multicolumn{1}{c|}{71.4}                  & 96.0                  & 81.0                 \\
\multicolumn{1}{l|}{MLDG  \cite{li2018learning}  }                             & \multicolumn{1}{c|}{79.5}                  & \multicolumn{1}{c|}{75.3}                  & \multicolumn{1}{c|}{71.5}                  & 94.3                  & 80.7                 \\
\multicolumn{1}{l|}{MASF \cite{dou2019domain}   }                             & \multicolumn{1}{c|}{80.3}                  & \multicolumn{1}{c|}{77.2}                  & \multicolumn{1}{c|}{71.7}                  & 95.0                  & 81.1                 \\
\multicolumn{1}{l|}{Epi-FCR  \cite{li2019episodic}  }                             & \multicolumn{1}{c|}{82.1}         & \multicolumn{1}{c|}{77.0}                  & \multicolumn{1}{c|}{\textbf{73.0}}                  & 93.9                  & 81.5                 \\
\multicolumn{1}{l|}{MMLD \cite{matsuura2020domain}  }                               & \multicolumn{1}{c|}{81.3}                  & \multicolumn{1}{c|}{77.2}                  & \multicolumn{1}{c|}{72.3}                  &    \textbf{96.1}               & 81.7                 \\
\rowcolor{gray!10} \multicolumn{1}{l|}{Ours  }                              & \multicolumn{1}{c|}{\textbf{82.2}}                  & \multicolumn{1}{c|}{\textbf{78.5}}         & \multicolumn{1}{c|}{72.5}         & 95.5       & \textbf{82.2}        \\ \hline
\end{tabular}}
\vspace{-1.0em}
\end{center}
\end{table}

In this section, we present more experiment results to further evaluate the effectiveness of our proposed DFF.

\subsection{More Experiments on the Task-1 (PACS)}
We further evaluate the effectiveness of our proposed \ours~on another commonly used dataset, \ieno, Office-Home \cite{venkateswara2017deep}, for investigating the domain generalization on the close-set classification. This dataset contains four domains (Aritistic, Clipart, Product and Real World) with 15,500 images of classes for home and office object recognition. Similar to the Protocol-1 on PACS dataset \cite{li2017deeper}, we adopt a ``Leave-One-Out'' protocol for the evaluation on Office-Home where three domains are used for training while the remaining one is for testing. The experiment results are shown in Table \ref{table:PACS}. Our proposed \ours~achieves significant improvements relative to the \emph{Baseline} model, and outperforms the state-of-the-art methods on this dataset by a clear margin over all evaluation settings. This further demonstrates the effectiveness of \ours.

\subsection{More Ablation Studies}

\begin{table}[t]

	\begin{center}

  \caption{Performance comparisons of different frequency transformations. In \textit{``Baseline"}, we take vanilla ResNet-18/-50 as the backbone models. \textit{``Wavelet (db3)"} and \textit{``Wavelet (Haar)"} denote the wavelet transforms with the Daubechies3 and Haar filters, respectively.}

  \label{table:frequency}
  \centering
  \scalebox{0.73}{\begin{tabular}{c|cccccc}
\hline
                                                                                                                             
                      \multirow{3}{*}{Method}   & \multicolumn{6}{c}{Source$\rightarrow$Target}                                                                                                                                                          \\  \cline{2-7} 
                        &    \multicolumn{2}{c|}{MS+CS+C3$\rightarrow$MA}                             & \multicolumn{2}{c|}{MS+MA+CS$\rightarrow$C3}                             & \multicolumn{2}{c}{MA+CS+C3$\rightarrow$MS}         \\ \cline{2-7} 
                        &\quad mAP           & \multicolumn{1}{c|}{R1}            & \quad mAP           & \multicolumn{1}{c|}{R1}            & \quad mAP           & R1            \\ \hline
 \multicolumn{1}{l|}{Base}                &  \quad 59.4          & \multicolumn{1}{c|}{83.1}         & \quad   30.3            & \multicolumn{1}{c|}{29.1}              & \quad 18.0              &   \multicolumn{1}{c}{41.9}     \\ \hline
 \multicolumn{1}{l|}{Wavelet (db3)}        &         \quad61.5& \multicolumn{1}{c|}{83.7}& \quad30.7& \multicolumn{1}{c|}{29.8}& \quad 18.3&42.2   \\    

 \multicolumn{1}{l|}{Wavelet (Haar)}                        & \quad61.1& \multicolumn{1}{c|}{ 83.6}& \quad30.5&\multicolumn{1}{c|}{ 29.7}& \quad18.5&42.3   \\      
\rowcolor{gray!10}  \multicolumn{1}{l|}{FFT (Ours)}               & \quad \textbf{71.1}         & \multicolumn{1}{c|}{\textbf{87.1}}         & \quad     \textbf{41.3}         & \multicolumn{1}{c|}{\textbf{41.1}}              & \quad        \textbf{25.1}       &   \multicolumn{1}{c}{\textbf{50.5}}     \\ \hline
\end{tabular}
  }

		\end{center}
	\vspace{-0.50cm}
\end{table}%

\begin{table}[t]

	\begin{center}

  \caption{Performance comparisons of different dimensions on which the Fast Fourier Transform (FFT) is performed. \textit{``FFT (CHW)"} refers to the models in which FFT is performed across the height (H), width (W) and channel (C) dimensions. In \textit{``FFT (HW)"}, we just perform FFT across the height and width dimensions, \ieno, for each feature map independently, which is the default setting in this paper.}

  \label{table:CHW}
  \centering
  \scalebox{0.73}{\begin{tabular}{c|cccccc}
\hline
                                                                                                                             
                      \multirow{3}{*}{Method}   & \multicolumn{6}{c}{Source$\rightarrow$Target}                                                                                                                                                          \\  \cline{2-7} 
                        &    \multicolumn{2}{c|}{MS+CS+C3$\rightarrow$MA}                             & \multicolumn{2}{c|}{MS+MA+CS$\rightarrow$C3}                             & \multicolumn{2}{c}{MA+CS+C3$\rightarrow$MS}         \\ \cline{2-7} 
                        &\quad mAP           & \multicolumn{1}{c|}{R1}            & \quad mAP           & \multicolumn{1}{c|}{R1}            & \quad mAP           & R1            \\ \hline
 \multicolumn{1}{l|}{Base}                &  \quad 59.4          & \multicolumn{1}{c|}{83.1}         & \quad   30.3            & \multicolumn{1}{c|}{29.1}              & \quad 18.0              &   \multicolumn{1}{c}{41.9}     \\ \hline
FFT (CHW)& \quad59.2&  \multicolumn{1}{c|}{83.0}& \quad30.0&  \multicolumn{1}{c|}{28.8} & \quad17.5&38.5\\ 

\rowcolor{gray!10}  \multicolumn{1}{l|}{FFT(HW)}               & \quad \textbf{71.1}         & \multicolumn{1}{c|}{\textbf{87.1}}         & \quad     \textbf{41.3}         & \multicolumn{1}{c|}{\textbf{41.1}}              & \quad        \textbf{25.1}       &   \multicolumn{1}{c}{\textbf{50.5}}     \\ \hline
\end{tabular}
  }

		\end{center}
	\vspace{-0.50cm}
\end{table}%

\paragraph{Experiments with other frequency transforms.}

We preliminarily investigate the effectiveness of using other frequency transforms in implementing our conceptualized \ours. In particular, we replace the Fast Fourier Transform (FFT) in our proposed scheme by the wavelet transform with two widely used filters, \ieno, db3 and Haar. From the experiment results in Table \ref{table:frequency}, we observe that adopting the wavelet transform also delivers improvements compared to \emph{Baseline}, but is inferior to adopting FFT. This is because the wavelet transform is a low frequency transformation such that our proposed filtering operation is performed in a local space, thus limiting the benefits of \ours.

\paragraph{Design choices of performing FFT.} 

In our proposed scheme, for the given intermediate feature $\mathbf{X}\in \mathbb{R}^{C \times H \times W}$, we perform FFT for each channel independently to obtain the latent frequency representations, as described in the Sec. 3.2 of the main paper. Here, we investigate other design choices of perform FFT. In the Table \ref{table:CHW}, we find that performing FFT across H, W, C dimensions leads to performance drop compared to \emph{Baseline}. For the intermediate feature $\mathbf{X}\in \mathbb{R}^{C \times H \times W}$, its different channels correspond to the outputs of different convolution kernels, which are independent in fact. Thus, we perform FFT on each channel of $\mathbf{X}$ independently.

\paragraph{Ablation study on the ratio $r$.}
We follow the overall architecture design of \cite{chi2020fast} to split the given intermediate feature $\mathbf{X}\in \mathbb{R}^{C \times H \times W}$ into $\mathbf{X}^{g} \in \mathbb{R}^{rC \times H \times W}$ and $\mathbf{X}^{l} \in \mathbb{R}^{(1-r)C \times H \times W}$ along its channel dimension with a ratio $r \in [0,1]$. Our proposed filtering operation is only performed on $\mathbf{X}^{g}$. When setting $r=0$, the models degenerate to the ResNet-18/-50 baselines. Setting $r=1$ means that we perform \ours~on the entire intermediate feature $\mathbf{X}$. As the experiment results in Table \ref{table:ratio}, we empirically find that the models with $r=0.5$ achieve the best performance, exploiting the complementarity of features in the frequency and original spaces.

\begin{figure}[t]
	\setlength{\abovecaptionskip}{0pt} 
\setlength{\belowcaptionskip}{-0pt}
	\begin{center}
		\includegraphics[width=1.08\linewidth]{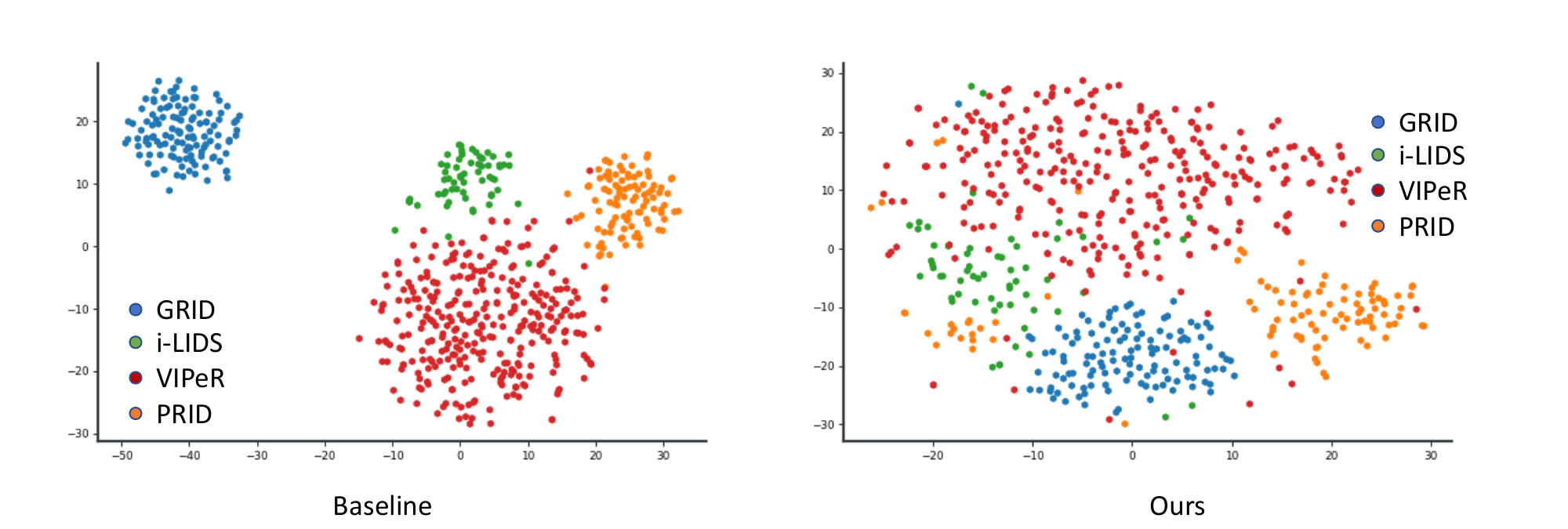}
	\end{center}
	\caption{The t-SNE \cite{van2008visualizing} visualization of ReID feature vectors learned by baseline (left) and our \ours~(right) on four unseen target datasets (GRID, i-LIDS, VIPeR and GRID). Best viewed in color.}
	\label{fig:TSNE}
	\vspace{-1.0em}
\end{figure}

\begin{table}[t]

	\begin{center}

  \caption{Performance comparisons of our proposed \ours~with different ratios. All models are built based on ResNet-18 for Task-1 while ResNet-50 for Task-2.}

  \label{table:ratio}
  \centering
  \scalebox{0.73}{\begin{tabular}{c|cccccc}
\hline
                                                                                                                             
                      \multirow{3}{*}{Ratio}   & \multicolumn{6}{c}{Source$\rightarrow$Target}                                                                                                                                                          \\  \cline{2-7} 
                        &    \multicolumn{2}{c|}{MS+CS+C3$\rightarrow$MA}                             & \multicolumn{2}{c|}{MS+MA+CS$\rightarrow$C3}                             & \multicolumn{2}{c}{MA+CS+C3$\rightarrow$MS}         \\ \cline{2-7} 
                        &\quad mAP           & \multicolumn{1}{c|}{R1}            & \quad mAP           & \multicolumn{1}{c|}{R1}            & \quad mAP           & R1            \\ \hline
 \multicolumn{1}{l|}{0.0}                &  \quad 59.4          & \multicolumn{1}{c|}{83.1}         & \quad   30.3            & \multicolumn{1}{c|}{29.1}              & \quad 18.0              &   \multicolumn{1}{c}{41.9}     \\ 
 \multicolumn{1}{l|}{0.25}                & \quad67.4& \multicolumn{1}{c|}{84.1 }& \quad38.1& \multicolumn{1}{c|}{38.1
 } & \quad22.9&48.4 \\   

 \multicolumn{1}{l|}{\textbf{0.5(Ours)}}                   & \quad \textbf{71.1}& \multicolumn{1}{c|}{\textbf{87.1}}& \quad\textbf{41.3}& \multicolumn{1}{c|}{\textbf{41.1}}& \quad\textbf{25.1}&\textbf{50.5}\\      
 \multicolumn{1}{l|}{0.75}  & \quad70.8& \multicolumn{1}{c|}{86.8} & \quad40.7& \multicolumn{1}{c|}{40.6}& \quad21.0&44.9\\
\multicolumn{1}{l|}{1.0}    &           \quad64.2& \multicolumn{1}{c|}{83.4}& \quad29.3& \multicolumn{1}{c|}{28.1}& \quad17.6&40.4\\ \hline
\end{tabular}
  }

		\end{center}
	\vspace{-0.50cm}
\end{table}%

\begin{table}[]
\caption{Performance comparisons of our proposed \ours~with the corresponding ResNet baselines on ImageNet-1K classification. ``\emph{DFF-ResNet-18/-50}'' denote the ResNet-18/-50 models equipped with our \ours.}
\label{table:imagenet}
\centering
\scalebox{0.73}{
\begin{tabular}{l|ccc}
\hline
{Method} & \quad Parameters \quad & \quad GFLOPs \quad & \quad Top-1 Acc.\quad \\ \hline
ResNet-18& 11.7M& 1.8& 69.8\\
\rowcolor{gray!10}DFF-ResNet-18& 12.2M& 2.0& 72.3\\ \hline
ResNet-50& 25.6M& 4.1& 76.3\\
\rowcolor{gray!10}DFF-ResNet-50& 27.7M& 4.5& 77.9\\ \hline
\end{tabular}}
\end{table}

\begin{table}[]
\caption{Performance comparisons of our proposed DFF with the state-of-the-art methods on supervised person ReID. ``\emph{Base.}'' refers to the baseline model.}
\label{table:supervised}
\centering
\scalebox{0.78}{
\begin{tabular}{l|cc|cc}
\hline
{\multirow{2}{*}{Model}} & \multicolumn{2}{c|}{Market-1501(MA)} & \multicolumn{2}{c}{MSMT17(MT)} \\ \cline{2-5} 
& \quad mAP& R1&\quad mAP& R1\\ \hline
PCB \cite{sun2018beyond} & 81.60 & 93.80 &-& - \\
BoT \cite{luo2019bag} & 85.90 & 94.50 &-& - \\
MGN \cite{wang2018learning} & 86.90 & 95.70 &-& - \\
JDGL \cite{zheng2019joint} & 86.00 & 94.80 & 52.30 & 77.20 \\
GASM \cite{he2020guided} & 84.70 & 95.30 & 52.50 & 79.50 \\
FPR \cite{he2019foreground} & 86.58 & 95.42 &-& - \\
HCTL \cite{zhao2020deep} & 81.80 & 93.80 & 43.60 & 74.30 \\
OSNet \cite{zhou2019omni} & 84.90 & 94.80 & 52.90 & 78.70 \\
RGA-SC \cite{zhang2020relation} & 88.40 & 96.10 & 57.50 & 80.30 \\
CDNet \cite{li2021combined} & 86.00 & 95.10 & 54.70 & 78.90 \\
Circle Loss \cite{sun2020circle} & 87.40 & 96.10 & 52.10 & 76.90 \\
AMD \cite{chen2021explainable} & 87.15 & 94.74 &-& - \\
FIDI \cite{yan2021beyond} & 86.80 & 94.50 &-& - \\
MPN-tuple \cite{zhangbeyond} & 88.70 & 95.30 & 60.10 & 82.20 \\ \hline
ResNet-50 Base. & 81.63 & 93.89 & 50.84 & 76.78 \\
\rowcolor{gray!10} DFF-ResNet-50& \textbf{90.21} & \textbf{96.17} & \textbf{60.21} & \textbf{82.95} \\ \hline
\end{tabular}}
\end{table}

\subsection{More Visualization Results}
We perform t-SNE visualization for the ReID feature vectors extracted by the baseline model and the model with our proposed \ours~on four unseen datasets. As shown in Fig.~\ref{fig:TSNE}, the four unseen target domains distribute more separately for the baseline model than that of ours. This indicates the domain gaps are effectively mitigated by our proposed \ourfull~(\ours).

\subsection{Effectiveness on ImageNet-1K Classification}
ImageNet-1K \cite{russakovsky2015imagenet} classification widely serves as a pre-training task, providing pre-trained weights as the model initialization for various downstream task. We present the effectiveness of our conceptualized \ours~on ImageNet-1K classification to showcase its potentials for more general purposes. As the results shown in Table \ref{table:imagenet}, our \ours~achieves 2.5\%/1.6\% improvements on the Top-1 classification accuracy compared to the corresponding baselines ResNet-18 and ResNet-50, respectively. Note that these improvements are achieved with the simple instantiation introduced in the Sec.3.3 of the main body. We believe more effective instantiations of \ours~are worth exploring to make \ours~contribute more in a wider range of fields.

\subsection{Effectiveness on Supervised Person ReID}
In the main body, we target domain generalization and present the effectiveness of our proposed \ours~on domain generalizable person ReID. In this supplementary material, we also showcase its potential on improving supervised person ReID. Following the previous works \cite{li2021combined,he2020guided,zheng2019joint,zhao2020deep,zhou2019omni} in this field, we evaluate our \ours~on two most widely used datasets Market-1501 \cite{zheng2015scalable} and MSMT17 \cite{wei2018person}. Note that another popular dataset DukeMTMC \cite{zheng2017unlabeled} has been taken down by its publisher. As shown in Table \ref{table:supervised}, the ResNet-50 equipped with \ours~significantly outperforms the baseline model and reaches the SOTA performance on this task. This demonstrates the proposed \ours~is also potentially beneficial for capturing discriminative features. We expect that it can contribute to more tasks.

\end{document}